\documentclass{article}

\usepackage[eandd, final]{neurips_2026}

\usepackage[utf8]{inputenc} 
\usepackage[T1]{fontenc}    
\usepackage{hyperref}       
\usepackage{url}            
\usepackage{booktabs}       
\usepackage{amsfonts}       
\usepackage{nicefrac}       
\usepackage{microtype}      
\usepackage{xcolor}         
\usepackage{graphicx}
\usepackage{algorithm}
\usepackage{algpseudocode}
\usepackage{amsmath}
\usepackage{multirow}
\usepackage[most]{tcolorbox}
\usepackage{amsthm}
\usepackage{afterpage}
\usepackage{subcaption}
\usepackage{xspace}

\definecolor{mygreen}{RGB}{0,100,0}
\definecolor{myred}{RGB}{180,30,30}

\newtcolorbox{principlebox1}[1][]{
  colback=blue!5!white,
  colframe=blue!60!black,
  fonttitle=\bfseries,
  title=Principle 1: Visual Discriminability,
  coltitle=black,
  boxed title style={colback=blue!15!white},
  sharp corners,
  enhanced,
  attach boxed title to top left={xshift=0.5cm, yshift=-2mm},
  #1
}

\newtcolorbox{principlebox2}[1][]{
  colback=blue!5!white,
  colframe=blue!60!black,
  fonttitle=\bfseries,
  title=Principle 2: Loop Closure,
  coltitle=black,
  boxed title style={colback=blue!15!white},
  sharp corners,
  enhanced,
  attach boxed title to top left={xshift=0.5cm, yshift=-2mm},
  #1
}

\newtcolorbox{principlebox3}[1][]{
  colback=blue!5!white,
  colframe=blue!60!black,
  fonttitle=\bfseries,
  title=Principle 3: Curriculum-based Progression,
  coltitle=black,
  boxed title style={colback=blue!15!white},
  sharp corners,
  enhanced,
  attach boxed title to top left={xshift=0.5cm, yshift=-2mm},
  #1
}

\title{LoopNav: Benchmarking Spatial Consistency in World Models}

\author{
  Kewei Lian\\
  NUS \\
  \And
  Shaofei Cai \\
  Peking University \\
  \And
  Yitao Liang \\ 
  Peking University \\
  \And
  Anji Liu\\
  NUS\\
}
\begin{document}
\newcommand{\loopnav}{\textsc{LoopNav}}
\newcommand{\matric}{Scene Graph Consistency Score\xspace}
\newcommand{\score}{\texttt{SGCS}\xspace}

\maketitle

\begin{abstract}
The ability to simulate the world in a spatially consistent manner is a crucial requirement for effective world models. Such a model enables high-quality visual generation, and also ensures the reliability of world models for downstream tasks such as simulation and planning. It must not only retain long-horizon observational information, but also enables the construction of explicit or implicit internal spatial representations. However, existing datasets do not explicitly enforce spatial consistency constraints, limiting both the ability to systematically evaluate this capability and to learn it through data-driven approaches. Furthermore, most existing benchmarks primarily emphasize visual coherence or generation quality, neglecting the requirement of long-range spatial consistency. To bridge this gap, we propose \loopnav, a dataset and corresponding benchmark centered on loop-based navigation for evaluating spatial consistency. The dataset comprises 200 hours of loop-based navigation videos with actions, collected from diverse locations in the open-world environment of Minecraft. We further introduce a \matric to quantify spatial consistency while remaining invariant to pixel-level variations. Dataset, benchmark, and code are open-sourced to support future research.

\end{abstract}
\afterpage{
\begin{figure}[ht]
    \centering 
    \includegraphics[width=\textwidth]{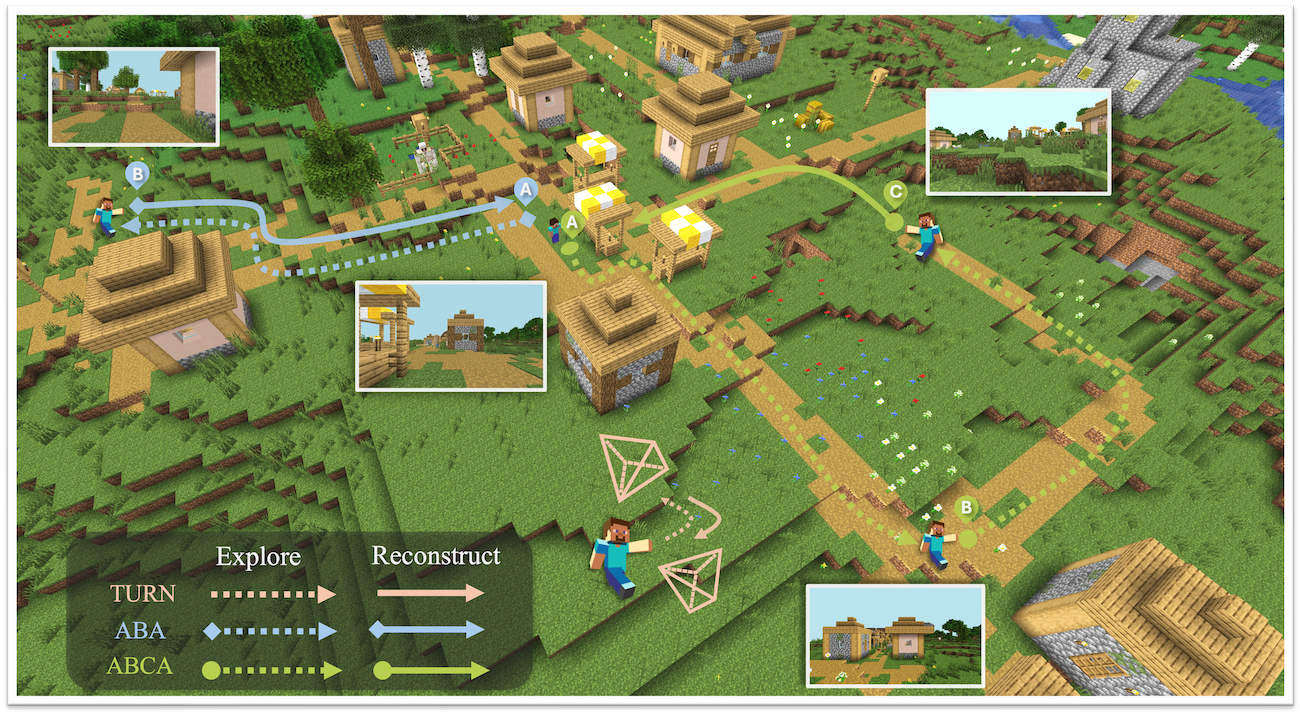}
    \vspace{-1.5em}
    \caption{\textbf{The Loop-Style Trajectory Data for Training and Benchmarking}. To be able to explicitly enforce spatial consistency, our dataset follows a loop-style navigation trajectory, including TURN, ABA and ABCA types. Our benchmark design explicitly separates the generation phase from the exploration phase and only evaluates the video quality of the generation part.}
    \label{fig:teaser}
\end{figure}
}
\section{Introduction}
Although recent advances in world models have significantly improved the visual quality of generated observations \citep{generalworldmodelsurvey,surveyding2024,sora,genie}, these models often struggle to preserve spatial consistency over extended rollouts \citep{oasis,guo2025mineworld}. Spatial consistency, the ability to preserve coherent and stable spatial structures across time, is essential for the reliability of world models in downstream applications such as model-based reinforcement learning \citep{bar2024navigationworldmodels,diamond,gameNGen}, autonomous driving \citep{DriveDreamer,vista}, and model predictive control \citep{du2023videolanguageplanning,dreamer,dreamerV3,unisim,yang2024videonewlanguagerealworld}. When violated, it can lead to severe failure modes: hallucinated structures, contradictions with prior observations, and visually incoherent scenes. For instance, in a world model guided navigation task, inconsistent rendering of revisited locations impedes global planning and loop closure. 

A key challenge underlying spatial inconsistency is the need to retain and utilize information over long temporal horizons. In principle, addressing this challenge requires models to maintain representations of previously observed spatial structures and re-anchor them upon revisitation. However, mainstream Transformer-based architectures are typically limited to short context windows due to the quadratic scaling of the attention mechanism \citep{vaswani2017attention}, making it difficult to directly incorporate long-range dependencies spanning hundreds or thousands of frames.

To mitigate this limitation, prior work has explored augmenting world models with memory mechanisms that enable long-term information retention \citep{graves2014neuralturingmachines}. These mechanisms may take various forms, including explicit spatial representations such as maps \citep{CMP,yang20253dmem3dscenememory}, coordinate systems \citep{wang2023gridmmgridmemorymap}, or graph-based structures \citep{savinov2018semiparametrictopo,kim2022topologicalsemanticgraphmemory}, as well as implicit representations based on learned embeddings or attention \citep{parisotto2017neuralmapstructuredmemory,cut3r}. While these approaches provide promising directions, they highlight a broader challenge: enabling models to maintain coherent spatial representations over long horizons remains an open problem.

Despite its importance, spatial consistency remains underexplored, particularly in terms of how it is supported, learned, and evaluated. A major reason for this gap is the absence of datasets that explicitly demand spatial consistency. Most existing datasets are designed for open-ended exploration, where agents continually encounter novel structures and objects while rarely revisiting previously seen locations. As a result, world models trained on such data tend to rely on global environmental priors to generate plausible observations, rather than leveraging episodic memory to reconstruct previously observed content. Meanwhile, evaluation metrics often prioritize visual fidelity and short-term temporal smoothness over long-term spatial coherence or logical consistency. This leads to models that function more as visually impressive but impractical ``dream machines'', limiting their applicability in downstream decision-making tasks.

Therefore, we argue that such a dataset must feature looped trajectories that revisit the same locations and objects from diverse viewpoints. We construct such a dataset, \loopnav, in the open-world environment of Minecraft. The dataset comprises over 200 hours of \textbf{loop-style} navigation trajectories across 147 diverse locations. Each trajectory follows a loop exploration pattern, ensuring that the same locations and layouts are revisited from varying camera views and times. This loop-based data structure naturally incentivizes models to learn long-horizon spatial consistency. Besides, we carefully select positions with diverse landmark objects to ensure that observations from different locations are \textbf{visually distinguishable}. To support progressive learning, the dataset also includes a \textbf{curriculum of sequence lengths}, enabling models to transition from short-term to long-term tasks. 

For benchmarking, we adopt an \textbf{explore-then-generate} protocol to evaluate spatial consistency. Specifically, we sample a loop trajectory (e.g., $A \rightarrow B \rightarrow A$), where the first segment ($A \rightarrow B$) is provided as context, and the remaining segment ($B \rightarrow A$) serves as the generation target. Existing evaluation metrics, such as SSIM and LPIPS \citep{lpips}, primarily assess visual quality. However, in world model generation, even when conditioned on identical action sequences, the generated trajectory may exhibit slight deviations—such as small translations, scaling, or accumulated drift—due to the inherent stochasticity of action execution and cumulative errors over time. These metrics can be sensitive to such misalignments, failing to reflect true spatial consistency.

To address this limitation, we introduce the \textbf{\matric (\score)}, an object-centric metric for evaluating spatial consistency. Specifically, video frames are first decomposed into category-wise object masks, and consistency is measured through object-level matching based on semantic category and spatial alignment. By comparing the matched object structures across trajectories, \score abstracts away pixel-level differences while remaining sensitive to semantic inconsistencies such as object disappearance, replacement, and large spatial layout changes.

In summary, our contributions are as follows:
\begin{itemize}
    \item \textbf{A dataset and benchmark for spatial consistency.}
    We introduce \loopnav, a dataset and benchmark centered on loop-based navigation, which bridges the gap between existing world model training setups and the need to evaluate long-horizon spatial consistency. It provides a unified platform for both training and evaluation under explicit spatial consistency constraints.

    \item \textbf{A structure-aware metric for spatial consistency.}
    We propose the \matric(\score), a structure-aware metric that evaluates spatial consistency while abstracting away pixel-level variations. It is robust to minor geometric misalignments, while remaining sensitive to semantic inconsistencies.

    \item \textbf{An open-source data collection framework.}
    We release a scalable and flexible pipeline for collecting navigation trajectories in Minecraft, which can be readily extended to other data collection settings. We also open-source the dataset and evaluation code to facilitate future research on spatial consistency in world models. Dataset, benchmark, and code are available
\footnote{Dataset page: \url{https://huggingface.co/datasets/kevinLian/LoopNav}}
\footnote{Code page: \url{https://github.com/Kevin-lkw/LoopNav}}.
\end{itemize}

\section{Related Work}
\subsection{World Models}
The goal of a world model is to simulate the environment: Given the current state and action, it predicts the next state and the corresponding reward. World models were originally proposed to improve sample efficiency in reinforcement learning \citep{oh2015actionconditionalvideopredictionusing,worldmodelHa,dreamer,dreamerV2,dreamerV3,TWM,diamond}. Different architectures, to varying degrees, retain information from the past to aid in future image prediction; however, they still lack an effective memory design capable of maintaining long-horizon spatial consistency.

Beyond model-based reinforcement learning, the potential of world models has also been increasingly explored in other domains such as game engines \citep{menapace2021playablevideogeneration,gameNGen,oasis}, autonomous driving \citep{hu2023gaia1generativeworldmodel,DriveDreamer,vista}, robot manipulation \citep{unisim,wu2022daydreamerworldmodelsphysical,bar2024navigationworldmodels}. In those downstream tasks, spatial consistency is even more important.

\subsection{Minecraft as an AI Testbed}
Minecraft is an open-world environment characterized by diverse terrains and rich interaction dynamics. Recent research has increasingly adopted Minecraft as a platform for training generative agents \citep{vpt,steve1,cai2023groot,cai2024groot2weaklysupervisedmultimodal,cai2025rocket1masteringopenworldinteraction,cai2025rocket2steeringvisuomotorpolicy} and constructing digital world models \citep{guo2025mineworld,oasis,hong2024slowfastvgenslowfastlearningactiondriven,song2025historyguidedvideodiffusion}. 

A more comprehensive review of related work is provided in the Appendix~\ref{app:E-related}. 

\section{\loopnav\ Dataset}
\begin{figure}
    \centering
    \includegraphics[width=\linewidth]{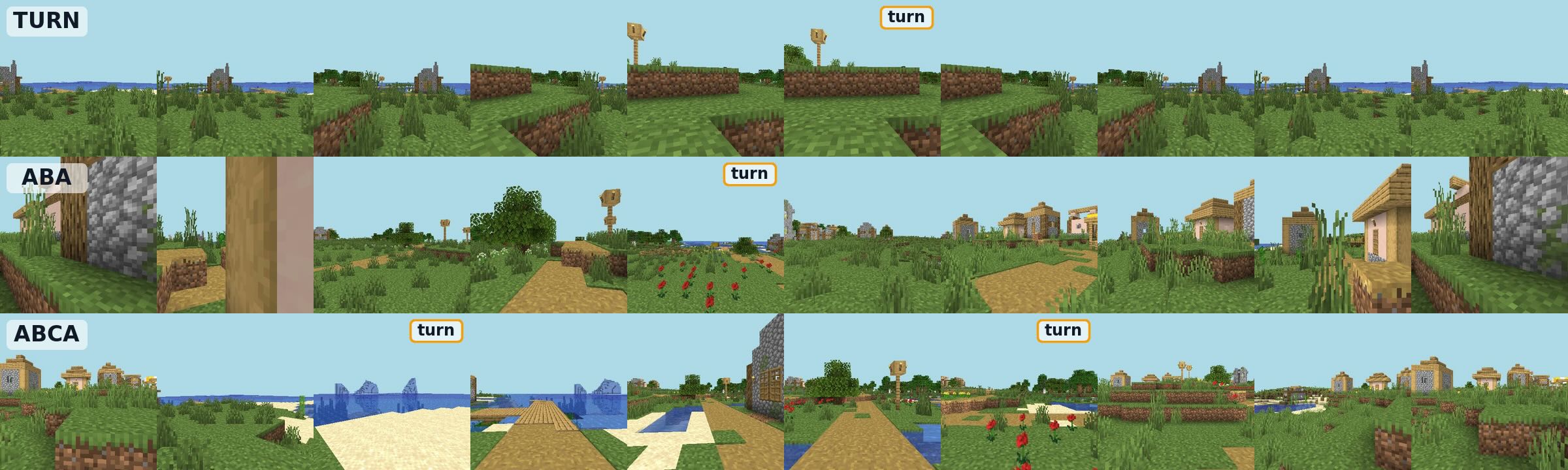}
    \caption{\textbf{Demostration of our dataset.} Rows show TURN, ABA , and ABCA , with time progressing from left to right. Ten frames are sampled from each trajectory, with turning points marked by orange turn labels.}
    \label{fig:placeholder}
\end{figure}

\subsection{Environment} 

We collect data in the open-world environment of Minecraft, which provides a suitable platform for large-scale and controlled data generation. Its procedurally generated worlds offer diverse and non-repetitive environments, while its mature ecosystem enables flexible customization and extension. Compared to real-world data collection, it also allows efficient, scalable, and highly parallelizable simulation. Further details are provided in Appendix~\ref{app:A1-mc}.

\subsection{Collection Pipeline}

We propose three principles for trajectory data collection: \textit{visual discriminability}, \textit{loop closure}, and \textit{curriculum-based progression}. Below, we detail each principle, followed by our implementation. 

\begin{principlebox1}
\textit{The sequence of observations along the trajectory should be visually distinguishable over time to capture meaningful scene variation. }
\end{principlebox1}

Within this procedurally generated world, we identify 6 categories comprising a total of 120 distinct villages, covering diverse terrains and architectural layouts. As players navigate through a village, visually distinctive structures such as houses, farms, and pathways serve as salient landmarks for spatial reasoning and cognitive map construction.  The ``visually distinguishable'' principle is particularly important for evaluating spatial consistency. If a scene lacks distinctive objects or structural variation, visually different states may become difficult to distinguish, making it ambiguous whether a generated frame is truly spatially consistent or merely visually similar. By ensuring sufficient variation in object appearance and spatial layout, our dataset enables more reliable evaluation of long-horizon spatial consistency. Detailed descriptions are provided in Appendix~\ref{app:B-dataset}.

\begin{principlebox2}
\textit{The generated trajectories must form at least one spatial loop, ensuring repeated visits to the same locations. }
\end{principlebox2}
We propose three types of trajectory structures: TURN, $A \rightarrow B \rightarrow A$ and $A \rightarrow B \rightarrow C \rightarrow A$, where A, B, and C denote distinct locations within the environment.

We use the ABA trajectory as a representative case to illustrate our data collection process. As detailed in Algorithm \ref{alg}, we first select a location $S$ from one of our collected sites,then randomly sample a starting point $A$ within the start range of $S$. This is done to ensure diversity in the starting positions within the same location. A target location $B$ is sampled within a specified navigation range. Trajectory recording begins: the agent performs a full 360-degree rotation at location $A$ to observe the surrounding environment and establish spatial context. It then navigates from $A$ to $B$ using the A* algorithm, and subsequently returns from $B$ to $A$, completing the loop. To reduce trajectory variance caused by independently planning the $A \to B$ and $B \to A$ paths with A*, we make sure the return path follows the recorded forward waypoints in reverse order. Details are provided in Appendix~\ref{app:B3stat}.





\begin{algorithm}[ht]
\caption{Loop Navigation Data Collection}
\label{alg}

\noindent
\begin{minipage}[t]{0.5\linewidth}
\vspace{0pt}
\begin{algorithmic}[1]
\Require Location $S$; ranges $R_{\mathrm{st}}$, $R_{\mathrm{nav}}$
\Ensure Loop trajectory $T$

\State $A \gets \textsc{SamplePoint}(S,R_{\mathrm{st}})$
\State $B \gets \textsc{SamplePoint}(A,R_{\mathrm{nav}})$
\State Teleport the agent to $A$ and rotate by $360^\circ$
\State $T_{AB} \gets \textsc{Navigate}(A,B)$
\State $T_{BA} \gets \textsc{Navigate}(B,A)$
\State \Return $\textsc{Concat}(T_{AB},T_{BA})$
\end{algorithmic}
\end{minipage}
\hfill
\begin{minipage}[t]{0.50\linewidth}
\vspace{0pt}
\begin{algorithmic}[1]
\Function{SamplePoint}{$C,r$}
    \Repeat
        \State $\Delta \sim \mathcal{U}([-r,r]^2)$
        \State $P \gets C+\Delta$
    \Until{$\|\Delta\|_2 \ge 0.8r
            \land \textsc{IsValid}(P)$}
    \State \Return $P$
\EndFunction
\end{algorithmic}
\end{minipage}

\end{algorithm}

\begin{principlebox3}
\textit{The trajectories follow a curriculum design, gradually expanding the spatial exploration radius to support progressive learning and evaluation. }
\end{principlebox3}
According to the exploration radius, we define four difficulty levels: $5$, $15$, $30$, and $50$ blocks. To improve training robustness and introduce variability, the target location $B$ is not sampled at a fixed distance. Instead, for a difficulty level $x$, we sample distances within the interval \([0.8x, \sqrt{2}x]\), ensuring diverse trajectory lengths while minimizing overlap between different difficulty levels. The full sampling procedure is also detailed in Algorithm \ref{alg}. Collect dataset trajectory length is visualized in Figure \ref{fig:frames}. 

\subsection{Implementation Details}
We collect 20 trajectories for each combination of location and exploration radius. The final dataset contains approximately 19.2 million frames, amounting to roughly 250 hours of navigation trajectories. 

We collect navigation trajectories in Minecraft using a customized data collection pipeline built on Mineflayer. 
Further details on the data collection system are provided in Appendix~\ref{app:A-setup}. Action design, and collected data format are in Appendix~\ref{app:B-dataset}.

\section{Benchmark}

\subsection{Loop-Based Evaluation Protocol}
Our benchmark is designed around $A \rightarrow B \rightarrow A$ loop trajectories. In this setting, we treat the $A \rightarrow B$ segment as an exploration phase, which serves as the contextual input to the model, and the $B \rightarrow A$ segment as a reconstruction phase, where the agent returns to a previously visited location. We provide an illustration of these phases in Figure \ref{fig:teaser}. As such, we evaluate model performance only on the $B \rightarrow A$ segment, as it provides a clear test of the model’s ability to maintain spatial consistency during long-horizon rollouts. Similarly, for longer $A \rightarrow B \rightarrow C \rightarrow A$ trajectories, we evaluate only the $C \rightarrow A$ segment. A detailed discussion is on Appendix~\ref{app:C3-ABCA}.

\begin{figure}
    \centering
    \includegraphics[width=\linewidth]{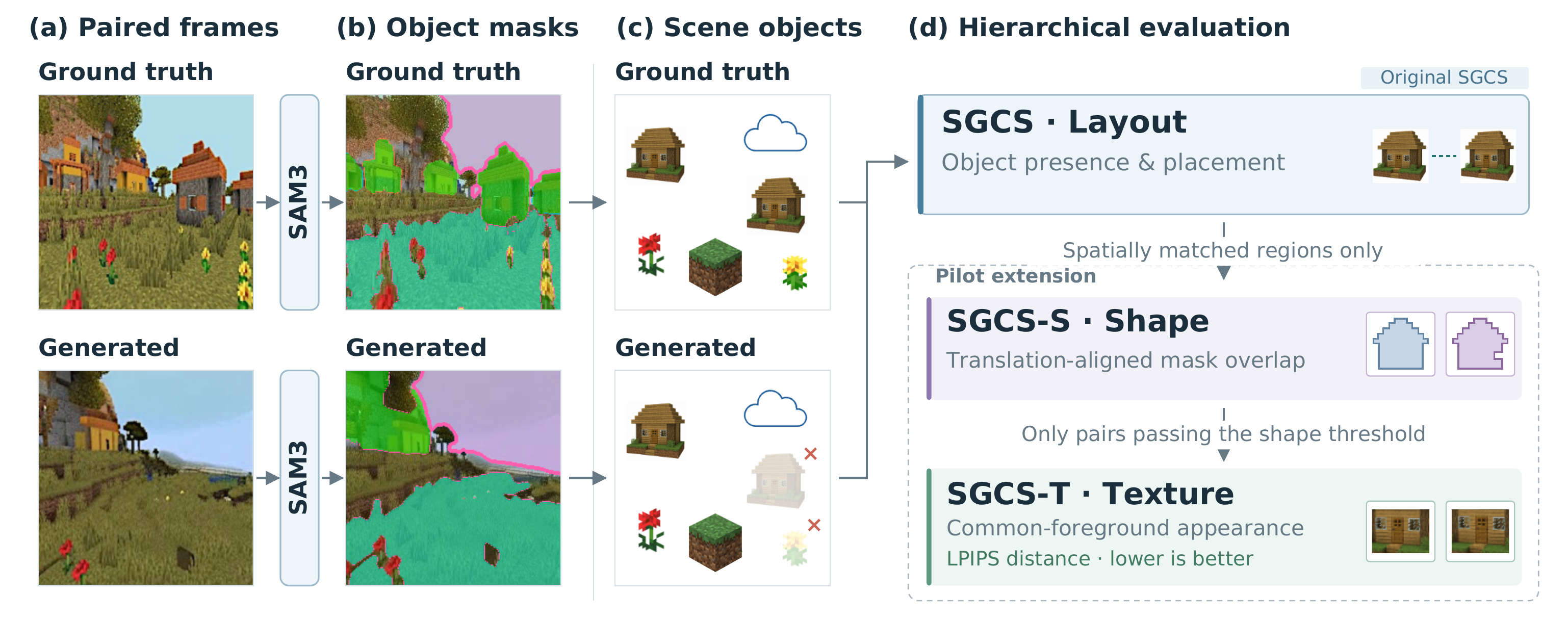}
    \caption{\textbf{Evaluation pipeline: \matric(\score).} We sample aligned frames from generated and ground-truth videos, extract category-wise object masks using SAM~3, and perform bipartite matching based on centroid distance. The final score aggregates category-level consistency with area weighting. We further introduce two pilot extensions: \score-Shape, which measures the centered (translation-normalized) mask IoU of layout-matched regions, and \score-Texture, which evaluates masked-crop LPIPS only for region pairs that pass the shape-consistency threshold.}
    \label{fig:benchmark}
\end{figure}

\subsection{Limitations of Visual Quality Metrics}

Given aligned ground-truth and generated trajectories, a natural approach is to evaluate their similarity using existing video generation metrics. However, such metrics are not well suited for assessing spatial consistency in world models. The generated trajectory may exhibit small geometric deviations due to accumulated errors in action execution. These deviations can manifest as slight translations, scaling, or viewpoint shifts over long rollouts. Importantly, such perturbations do not necessarily indicate a failure of spatial consistency. However, visual quality metrics are highly sensitive to these geometric perturbations. They may penalize predictions that are semantically consistent but not perfectly aligned with the ground truth, leading to low scores. At the same time, these metrics fail to capture spatial consistency itself. They primarily assess visual fidelity or perceptual similarity, rather than whether the generated trajectory remains consistent with previously observed scenes. Consequently, a visually plausible but spatially inconsistent generation may still receive a high score.

\subsection{\matric}

To address the limitations of pixel-level metrics, we introduce the \textbf{\matric (\score)}. \score is designed to be robust to minor geometric perturbations, such as translation, rotation, and scaling, while remaining sensitive to changes in spatial consistency, including object disappearance, replacement, and incorrect spatial relationships.

Figure~\ref{fig:benchmark} illustrates our evaluation pipeline. Given a generated video and its corresponding ground-truth video of equal length, we first sample aligned frame pairs at a fixed temporal stride. For each pair of frames, we compute \score based on category-wise object matching.

Specifically, for each semantic category $c$ (e.g., building, tree, flower, grass), we segment both frames into object masks using SAM~3 \citep{sam3}. We compute the centroid of each mask and identify matchable pairs based on their normalized spatial distance. Intuitively, two masks are considered a valid match if their centers are sufficiently close relative to the image size.

Since multiple instances of the same object category may appear in a frame, we perform bipartite matching (via the Hungarian algorithm \citep{hungarian}) to obtain the maximum number of matched object pairs. Based on the number of matched and unmatched masks, we compute a category-level consistency score that reflects how well objects of category $c$ are preserved. The final frame-level \score is obtained as the weighted average over all categories. For video-level evaluation, we compute \score for all sampled frame pairs and report their average. Importantly, \score abstracts away pixel-level differences and instead evaluates consistency at the level of object presence and spatial arrangement. As a result, it is robust to minor geometric misalignments, while remaining sensitive to spatial inconsistencies. We leave the details of implimentation in Appendix \ref{app:C1-SGCS}.

\subsection{Extension of \matric}

While the original \score evaluates whether the location objects of the same semantic category align across frames, spatial location captures only coarse layout consistency. To provide a fine-grained evaluation, we introduce \score-Shape and \score-Texture, which measure object silhouette fidelity and surface appearance, respectively. These metrics establish a natural coarse-to-fine hierarchy: comparing object masks (\score-Shape) is meaningful only when coarse spatial positions are aligned (high \score); similarly, comparing visual textures (\score-Texture) is conditioned on achieving high shape consistency. We conducted preliminary experiments to validate the sensitivity and orthogonality of this hierarchical extension in Appendix \ref{app:C4-sgcsextend}.

\textbf{\score-Shape.} For each layout-matched pair, we align the two masks by their centroids and compute mask \textbf{IoU} without resizing. This removes translation while preserving differences in object size and silhouette. 

\textbf{\score-Texture.} For pairs passing a shape-consistency threshold, we compare their aligned foreground crops using LPIPS after replacing the background with a neutral color. Lower LPIPS indicates more consistent appearance, and we jointly report the coverage of eligible pairs.

\section{Experiments}
\subsection{\score Metric Validation}
\begin{figure}[t]
    \centering

    \begin{subfigure}[t]{0.53\linewidth}
        \centering
        \includegraphics[width=\linewidth]{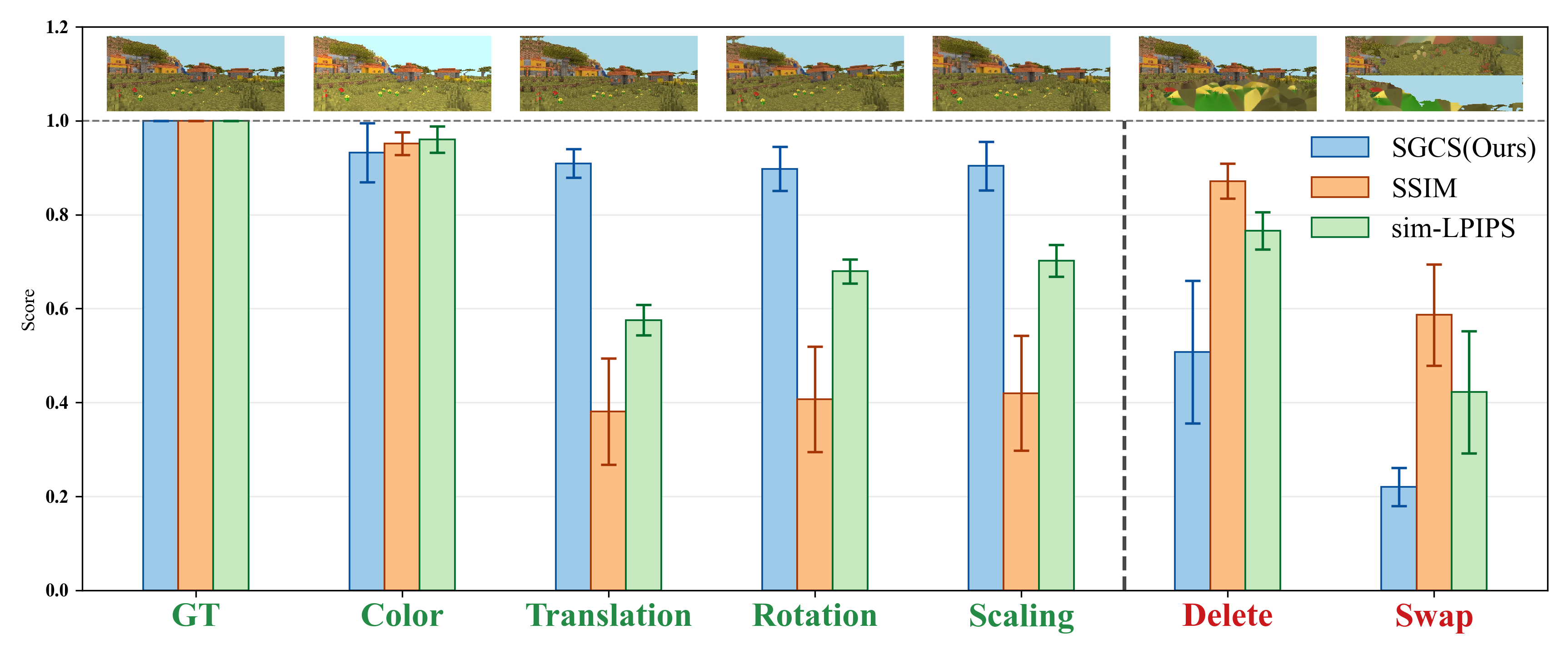}
        \caption{Controlled perturbation comparison.}
        \label{fig:sgcs_sense}
    \end{subfigure}
    \begin{subfigure}[t]{0.46\linewidth}
        \centering
        \includegraphics[width=\linewidth]{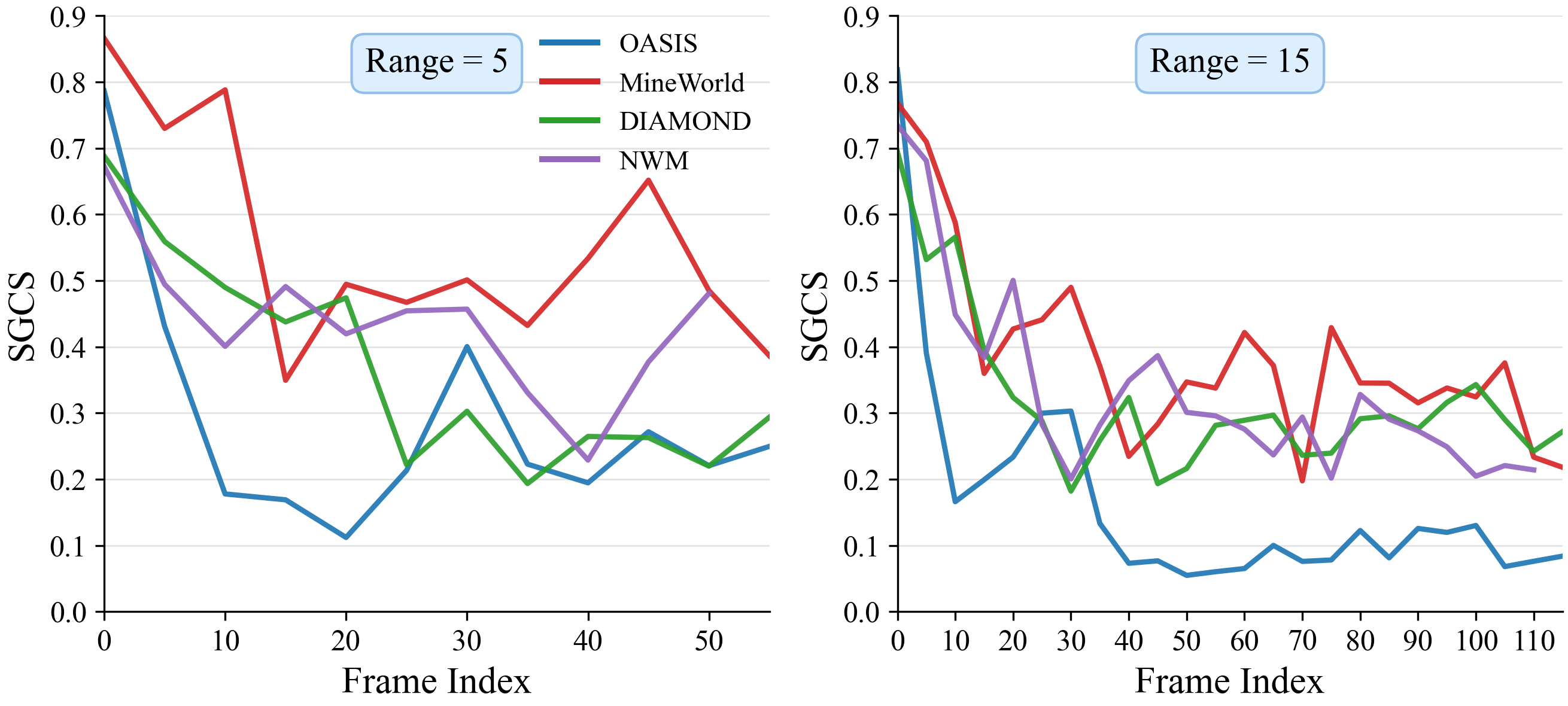}
        \caption{Per-frame \score over time.}
        \label{fig:per_frame_sgcs}
    \end{subfigure}

    \caption{\textbf{Left: Comparison of metrics under controlled perturbations.} From left to right, \textcolor{mygreen}{Geometric Perturbations}: ground truth, color change, small translation, small rotation, and scaling; \textcolor{myred}{Spatial Inconsistencies}: object deletion and object position swapping. We compare \score (ours), SSIM, and sim-LPIPS. Higher values indicate better.
    \textbf{Right: Per-frame \score.} Spatial consistency degrades across all baselines as generation progresses.}
    \label{fig:exp_plot}
\end{figure}

We evaluate \score under controlled perturbations to validate its robustness and sensitivity. Specifically, we apply geometric perturbations (colour change, translation, rotation, scaling) and semantic modifications (object deletion and swapping) to ground-truth videos. We compare \score, SSIM, and sim-LPIPS (defined as 1-LPIPS). All metrics are normalized such that higher values indicate better consistency and GT achieves a score of 1. As shown in Figure \ref{fig:sgcs_sense}, \score remains stable under geometric perturbations, demonstrating robustness to pixel-level misalignment. In contrast, SSIM and LPIPS degrade significantly under some transformations.

Conversely, under semantic modifications, \score decreases sharply, correctly identifying violations of spatial consistency, while SSIM and LPIPS may still assign relatively high scores due to preserved visual appearance.

These results confirm that \score achieves the desired balance between invariance to geometric perturbations and sensitivity to spatial inconsistencies. We leave more demonstrations of synthetic perturbation on Figure \ref{fig:synthetic} and the details of implementation on Appendix \ref{app:D0-synthetic}.

\subsection{World Model Baselines}
\begin{figure}[ht!]
    \centering 
    \includegraphics[width=\textwidth]{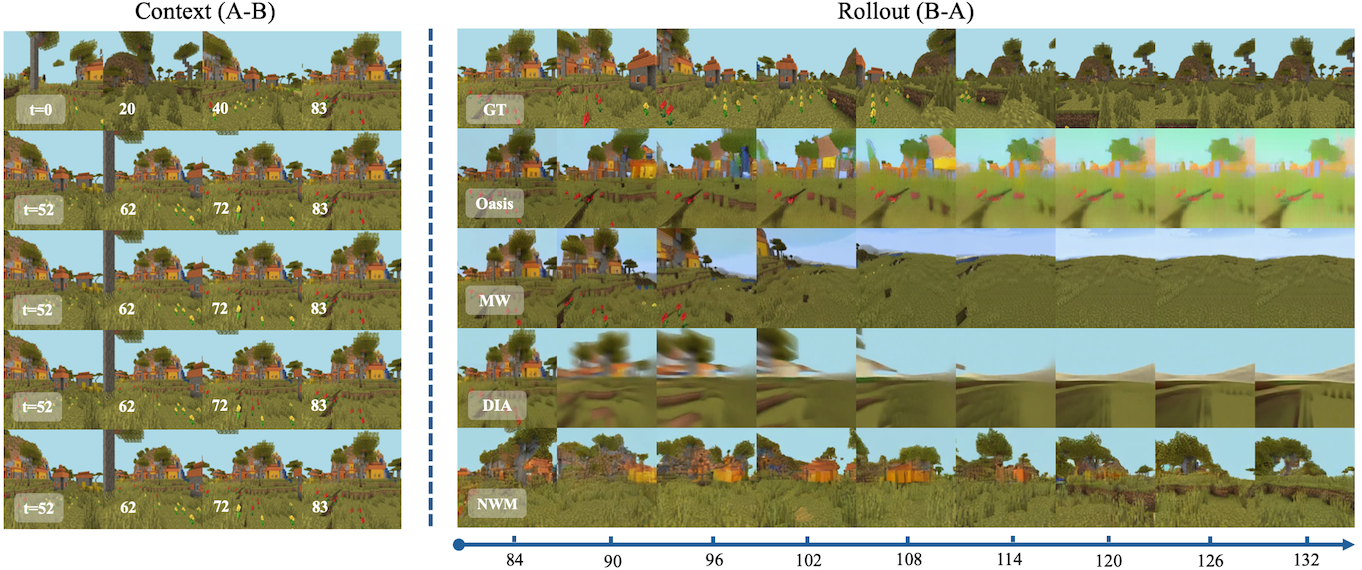}
    \vspace{-0.5em}
    \caption{\textbf{Qualitative Result of Four Baselines}. Top to buttom: Ground Truth(GT), Oasis, Mineworld(MW), DIAMOND(DIA), Navigation World Model(NWM). Leftmost label “t=52” indicates the start context range accepted by each model. All models begin rollout from frame 84.}
    \label{fig:res}
\end{figure}
We evaluate the following baselines on our dataset and benchmark. We set the context length of all baselines except WorldMem to 32 frames.

\textbf{Oasis} \citep{oasis} is a world model with a ViT \citep{ViT}  as spatial autoencoder, and a DiT \citep{DiT}  as latent diffusion backbone. We use the pretrained checkpoint.
\textbf{Mineworld} \citep{guo2025mineworld} is another interactive world model driven by a visual-action autoregressive Transformer pretrained on Minecraft. We use the pretrained checkpoint.
\textbf{DIAMOND} \citep{diamond} is a diffusion-based world model built upon the UNet architecture \citep{unet}.  We pretrain a model using our dataset from scratch. 
\textbf{Navigation World Model} 
\citep{bar2024navigationworldmodels} is a controllable video generation model that predicts future visual observations conditioned on past observations and navigation actions. We pretrain NWM model using our dataset from scratch. 
\textbf{WorldMem} \citep{xiao2026worldmemlongtermconsistentworld} is a memory-augmented diffusion world model that retrieves spatially relevant past views, we fine-tune its action and pose conditioning together with lightweight attention adapters.
The implementation details and additional results are provided in the Appendix \ref{app:D-experiment}.

\subsection{Experiment Settings}
\label{expsetting}
Although our dataset includes data from villages, biomes, and structures, we focus solely on village environments for training and evaluation. This decision is based on the large number of villages in Minecraft and the inherent diversity of them. We use village with id 1-16 as training set, 17-18 as validation set and 19-20 as test set. This is because during training, the layout of a village can be memorized by model.

\textbf{Evaluation} We evaluate each model across all navigation ranges. Due to the large scale of the test dataset, we do not perform evaluation on the entire set. Instead, we sample the 5 trajectories from each of the 6 villages with index 19 and 20, resulting in 60 evaluation trajectories per navigation range. Each trajectory is assessed using the metrics described in the benchmark section. We report the average performance across all 60 trajectories, as shown in Table~\ref{table}. In addition, we randomly select one trajectory to present qualitative results, shown in Figure~\ref{fig:res}.

\subsection{Analysis}
\begin{table*}[t]
    \centering
    \caption{
        SGCS ($\uparrow$) across TURN, ABA, and ABCA.
        Results are reported as mean $\pm$ standard deviation over $60$ trajectories.
        Bold means indicate the best results.
    }
    \label{tab:sgcs_by_range}

    \begingroup
    \small
    \setlength{\tabcolsep}{4pt}
    \renewcommand{\arraystretch}{1.08}

    \newcommand{\sgcs}[2]{%
        \ensuremath{#1{\scriptstyle\,\pm\,#2}}%
    }
    \newcommand{\bestsgcs}[2]{%
        \ensuremath{\mathbf{#1}{\scriptstyle\,\pm\,#2}}%
    }

    \begin{tabular}{@{}ccccccc@{}}
        \toprule
        Task & Range & Oasis & NWM & MineWorld & DIAMOND & WorldMem \\
        \midrule

        \multirow{4}{*}{TURN}
        & $60^\circ$--$90^\circ$
        & \sgcs{0.555}{0.21}
        & \sgcs{0.344}{0.23}
        & \sgcs{0.721}{0.17}
        & \sgcs{0.597}{0.23}
        & \bestsgcs{0.839}{0.14} \\

        & $90^\circ$--$120^\circ$
        & \sgcs{0.458}{0.22}
        & \sgcs{0.293}{0.19}
        & \sgcs{0.665}{0.21}
        & \sgcs{0.484}{0.23}
        & \bestsgcs{0.846}{0.12} \\

        & $120^\circ$--$150^\circ$
        & \sgcs{0.472}{0.24}
        & \sgcs{0.291}{0.20}
        & \sgcs{0.597}{0.20}
        & \sgcs{0.470}{0.26}
        & \bestsgcs{0.839}{0.13} \\

        & $150^\circ$--$180^\circ$
        & \sgcs{0.458}{0.20}
        & \sgcs{0.253}{0.16}
        & \sgcs{0.593}{0.18}
        & \sgcs{0.510}{0.23}
        & \bestsgcs{0.829}{0.11} \\

        \midrule

        \multirow{4}{*}{ABA}
        & $5$
        & \sgcs{0.308}{0.24}
        & \sgcs{0.398}{0.26}
        & \sgcs{0.413}{0.24}
        & \sgcs{0.311}{0.26}
        & \bestsgcs{0.530}{0.25} \\

        & $15$
        & \sgcs{0.263}{0.20}
        & \bestsgcs{0.382}{0.24}
        & \sgcs{0.367}{0.23}
        & \sgcs{0.306}{0.23}
        & \sgcs{0.373}{0.21} \\

        & $30$
        & \sgcs{0.184}{0.15}
        & \bestsgcs{0.376}{0.21}
        & \sgcs{0.295}{0.24}
        & \sgcs{0.198}{0.19}
        & \sgcs{0.195}{0.15} \\

        & $50$
        & \sgcs{0.110}{0.09}
        & \bestsgcs{0.352}{0.21}
        & \sgcs{0.325}{0.23}
        & \sgcs{0.181}{0.18}
        & \sgcs{0.108}{0.07} \\

        \midrule

        \multirow{4}{*}{ABCA}
        & $5$
        & \sgcs{0.352}{0.28}
        & \sgcs{0.390}{0.30}
        & \sgcs{0.451}{0.26}
        & \sgcs{0.349}{0.26}
        & \bestsgcs{0.551}{0.27} \\

        & $15$
        & \sgcs{0.273}{0.22}
        & \sgcs{0.402}{0.25}
        & \bestsgcs{0.421}{0.24}
        & \sgcs{0.276}{0.22}
        & \sgcs{0.415}{0.25} \\

        & $30$
        & \sgcs{0.205}{0.21}
        & \bestsgcs{0.365}{0.25}
        & \sgcs{0.330}{0.23}
        & \sgcs{0.237}{0.23}
        & \sgcs{0.291}{0.25} \\

        & $50$
        & \sgcs{0.204}{0.21}
        & \bestsgcs{0.391}{0.22}
        & \sgcs{0.376}{0.26}
        & \sgcs{0.218}{0.23}
        & \sgcs{0.329}{0.27} \\

        \bottomrule
    \end{tabular}

    \endgroup
\end{table*}
\textbf{Overall Performance}
The evaluated models exhibit substantial limitations in maintaining spatial consistency over long rollouts. As illustrated in Figure~\ref{fig:per_frame_sgcs} and Figure~\ref{fig:res}, initially plausible reconstructions can gradually diverge from the previously observed environment.
Importantly, spatial inconsistency does not always coincide with a collapse in visual quality: a model may continue to generate plausible scenes while failing to preserve the structures encountered during exploration.
As shown in Table \ref{tab:sgcs_by_range}, WorldMem achieves the highest \score on TURN and short-range navigation, demonstrating the potential of memory-augmented generation for revisitation.
However, its advantage diminishes on longer navigation trajectories, just as oasis(Worldmem is finetuned from the base model of Oasis), while NWM achieves the highest scores in several longer-range settings.

\textbf{Metric Fidelity to Model Performance and Task Difficulty} As shown in Table~\ref{table}, our metric exhibits strong correlation with perceived visual quality. The ranking is consistent with human qualitative observations. Furthermore, the results reveal a clear difficulty gradient in our task design: as the navigation range increases, the task becomes more challenging, leading to a consistent decrease in \score. 

In contrast, traditional visual quality metrics, including SSIM, LPIPS, and FVD, fail to faithfully reflect the relative performance of the baseline models and do not capture the increasing task difficulty. These observations are consistent with our earlier findings, highlighting the limitations of conventional visual similarity metrics in evaluating spatial consistency.

\textbf{Outlier Analysis} We observe a single outlier where MineWorld and NWM achieve higher performance at ABCA type navgation, trajectory length 50 compared to length 30. This occurs because, at length 50, the task corresponds to extremely long-horizon navigation (approximately 1200 frames; see Fig.~\ref{fig:frames}). Additionaly, ABCA trajectories involve more extensive exploration and wandering, often leading the agent to move beyond the village. In such scenarios, many frames encountered during navigation consist primarily of simple backgrounds, such as grass and sky.  As a result, visually discriminative objects may become sparse or absent.  On the other hand, even when the model fails to maintain spatial consistency, it can still generate visually similar base scenes (i.e., combinations of grass and sky; see later frames in Fig.~\ref{fig:res}). Consequently, this leads to artificially higher consistency scores.

This exception does not affect our overall conclusions. Rather, it further underscores the importance of the ``visually distinguishable'' principle in dataset construction. At the same time, it points to a promising direction for future work, such as incorporating different weighting across trajectory segments to better account for varying scene complexity.

\begin{table}[ht]
\footnotesize
\setlength{\tabcolsep}{1.35 mm}
\centering
\caption{\textbf{Evaluation Results of Four Baselines under Different Navigation Ranges}. The \includegraphics[width=0.017\linewidth]{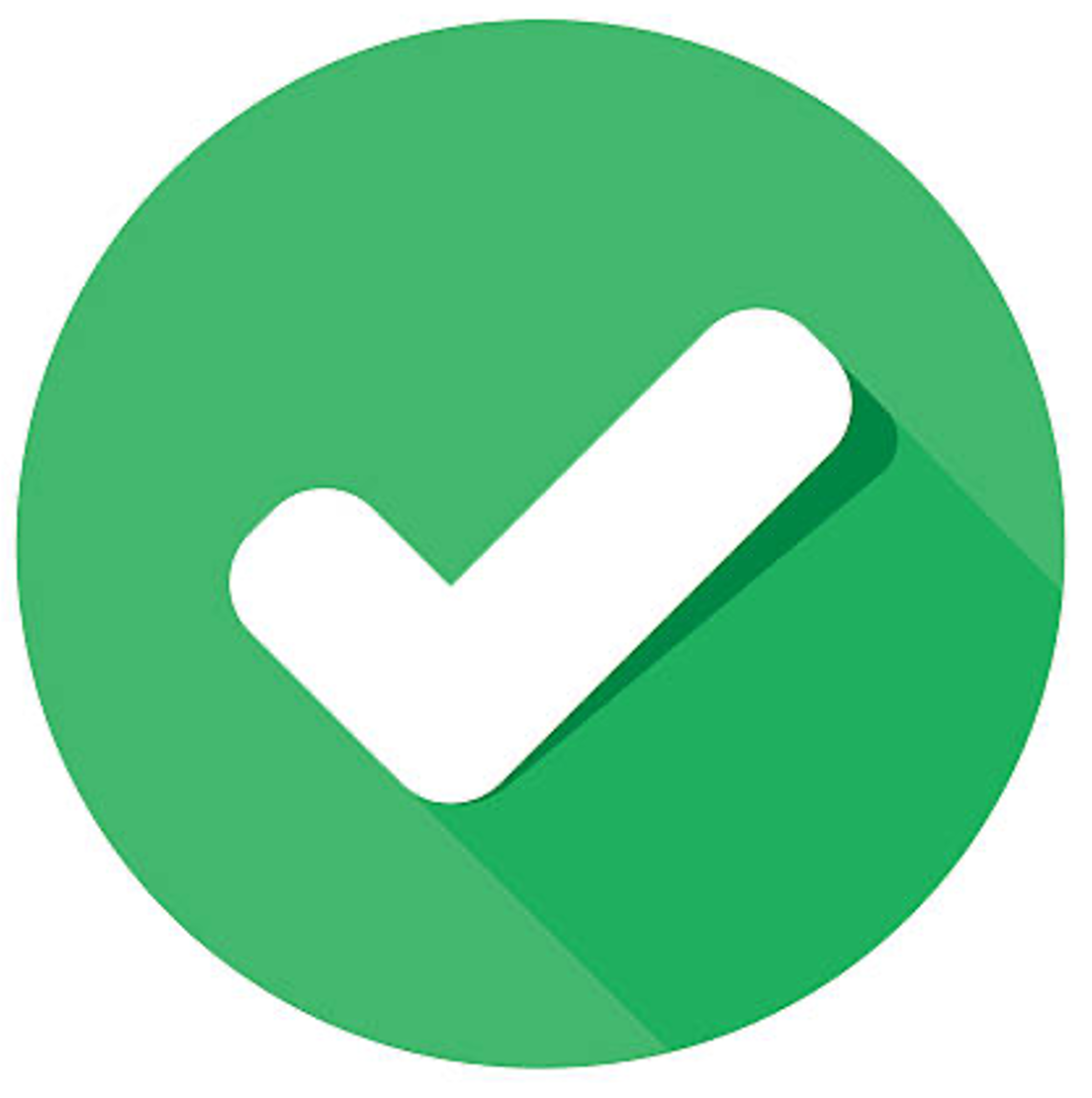} icon indicates that the model was trained on our dataset, while \includegraphics[width=0.017\linewidth]{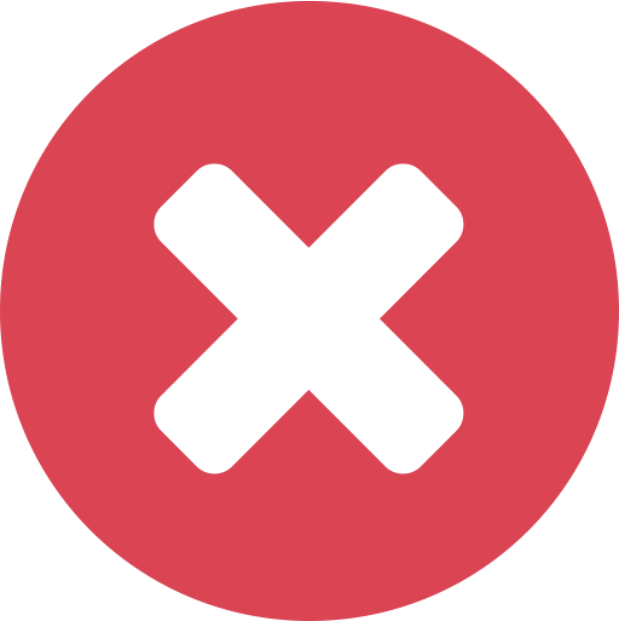} indicates using a pretrained checkpoint. ABA and ABCA denotes the navigation type, and the suffix of model(-5,-15,-30,-50) denotes the navigation range.}
\begin{tabular}{l@{}c@{\hspace{1mm}}c@{\hspace{1mm}}cc@{\hspace{1mm}}cc@{\hspace{1mm}}cc@{\hspace{1mm}}c}
\toprule
\multirow{2}{*}{\textbf{Model}} & \multirow{2}{*}{\textbf{Train}} 
& \multicolumn{2}{c}{\textbf{SGCS $\uparrow$}}
& \multicolumn{2}{c}{\textbf{SSIM $\uparrow$}} 
& \multicolumn{2}{c}{\textbf{LPIPS $\downarrow$}} 
& \multicolumn{2}{c}{\textbf{FVD $\downarrow$}} \\
\cmidrule(r){3-4} \cmidrule(r){5-6} \cmidrule(r){7-8} \cmidrule(r){9-10}
 & & ABA & ABCA & ABA & ABCA & ABA & ABCA & ABA & ABCA \\
\midrule

Oasis-5  & \includegraphics[width=0.02\linewidth]{contents/cross.png} 
         & \(\mathbf{0.31} {\scriptstyle \pm 0.24}\) & \(\mathbf{0.35} {\scriptstyle \pm 0.28}\)
         & \(0.36 {\scriptstyle \pm 0.13}\) & \(0.34 {\scriptstyle \pm 0.12}\) 
         & \(0.76 {\scriptstyle \pm 0.09}\) & \(0.82 {\scriptstyle \pm 0.11}\) 
         & \(2615 {\scriptstyle \pm 1067}\) & \(2583 {\scriptstyle \pm 647}\) \\

Oasis-15 & \includegraphics[width=0.02\linewidth]{contents/cross.png} 
         & \(\mathbf{0.26} {\scriptstyle \pm 0.20}\) & \(\mathbf{0.27} {\scriptstyle \pm 0.22}\)
         & \(0.37 {\scriptstyle \pm 0.12}\) & \(0.38 {\scriptstyle \pm 0.14}\) 
         & \(0.82 {\scriptstyle \pm 0.08}\) & \(0.81 {\scriptstyle \pm 0.10}\) 
         & \(2516 {\scriptstyle \pm 567}\) & \(3146 {\scriptstyle \pm 1055}\) \\

Oasis-30 & \includegraphics[width=0.02\linewidth]{contents/cross.png} 
         & \(\mathbf{0.18} {\scriptstyle \pm 0.15}\) & \(\mathbf{0.21} {\scriptstyle \pm 0.21}\)
         & \(0.33 {\scriptstyle \pm 0.11}\) & \(0.35 {\scriptstyle \pm 0.11}\) 
         & \(0.86 {\scriptstyle \pm 0.08}\) & \(0.85 {\scriptstyle \pm 0.09}\) 
         & \(3131 {\scriptstyle \pm 713}\) & \(3199 {\scriptstyle \pm 1000}\) \\

Oasis-50 & \includegraphics[width=0.02\linewidth]{contents/cross.png} 
         & \(\mathbf{0.11} {\scriptstyle \pm 0.09}\) & \(\mathbf{0.20} {\scriptstyle \pm 0.21}\)
         & \(0.36 {\scriptstyle \pm 0.12}\) & \(0.36 {\scriptstyle \pm 0.11}\) 
         & \(0.86 {\scriptstyle \pm 0.09}\) & \(0.83 {\scriptstyle \pm 0.07}\) 
         & \(3334 {\scriptstyle \pm 658}\) & \(3162 {\scriptstyle \pm 1245}\) \\

\midrule

Mineworld-5  & \includegraphics[width=0.02\linewidth]{contents/cross.png} 
             & \(\mathbf{0.41} {\scriptstyle \pm 0.24}\) & \(\mathbf{0.45} {\scriptstyle \pm 0.26}\)
             & \(0.31 {\scriptstyle \pm 0.09}\) & \(0.32 {\scriptstyle \pm 0.10}\) 
             & \(0.73 {\scriptstyle \pm 0.05}\) & \(0.72 {\scriptstyle \pm 0.07}\) 
             & \(2089 {\scriptstyle \pm 1007}\) & \(1914 {\scriptstyle \pm 660}\) \\

Mineworld-15 & \includegraphics[width=0.02\linewidth]{contents/cross.png} 
             & \(\mathbf{0.37} {\scriptstyle \pm 0.23}\) & \(\mathbf{0.42} {\scriptstyle \pm 0.24}\)
             & \(0.34 {\scriptstyle \pm 0.13}\) & \(0.32 {\scriptstyle \pm 0.11}\) 
             & \(0.74 {\scriptstyle \pm 0.08}\) & \(0.74 {\scriptstyle \pm 0.07}\) 
             & \(2367 {\scriptstyle \pm 770}\) & \(2009 {\scriptstyle \pm 921}\) \\

Mineworld-30 & \includegraphics[width=0.02\linewidth]{contents/cross.png} 
             & \(\mathbf{0.30} {\scriptstyle \pm 0.24}\) & \(\mathbf{0.33} {\scriptstyle \pm 0.23}\)
             & \(0.33 {\scriptstyle \pm 0.13}\) & \(0.28 {\scriptstyle \pm 0.09}\) 
             & \(0.77 {\scriptstyle \pm 0.08}\) & \(0.77 {\scriptstyle \pm 0.08}\) 
             & \(2316 {\scriptstyle \pm 945}\) & \(2094 {\scriptstyle \pm 1047}\) \\

Mineworld-50 & \includegraphics[width=0.02\linewidth]{contents/cross.png} 
             & \(\mathbf{0.32} {\scriptstyle \pm 0.23}\) & \(\mathbf{0.38} {\scriptstyle \pm 0.26}\)
             & \(0.31 {\scriptstyle \pm 0.16}\) & \(0.32 {\scriptstyle \pm 0.12}\) 
             & \(0.78 {\scriptstyle \pm 0.12}\) & \(0.75 {\scriptstyle \pm 0.10}\) 
             & \(2077 {\scriptstyle \pm 632}\) & \(2144 {\scriptstyle \pm 898}\) \\

\midrule

DIAMOND-5  & \includegraphics[width=0.02\linewidth]{contents/tick.png}  
           & \(\mathbf{0.31} {\scriptstyle \pm 0.26}\) & \(\mathbf{0.35} {\scriptstyle \pm 0.26}\)
           & \(0.40 {\scriptstyle \pm 0.10}\) & \(0.37 {\scriptstyle \pm 0.09}\) 
           & \(0.75 {\scriptstyle \pm 0.09}\) & \(0.79 {\scriptstyle \pm 0.09}\) 
           & \(3353 {\scriptstyle \pm 1242}\) & \(3336 {\scriptstyle \pm 1392}\) \\

DIAMOND-15 & \includegraphics[width=0.02\linewidth]{contents/tick.png}  
           & \(\mathbf{0.31} {\scriptstyle \pm 0.23}\) & \(\mathbf{0.28} {\scriptstyle \pm 0.22}\)
           & \(0.38 {\scriptstyle \pm 0.10}\) & \(0.39 {\scriptstyle \pm 0.10}\) 
           & \(0.78 {\scriptstyle \pm 0.08}\) & \(0.79 {\scriptstyle \pm 0.09}\) 
           & \(3691 {\scriptstyle \pm 937}\)  & \(3302 {\scriptstyle \pm 1191}\) \\

DIAMOND-30 & \includegraphics[width=0.02\linewidth]{contents/tick.png}  
           & \(\mathbf{0.20} {\scriptstyle \pm 0.19}\) & \(\mathbf{0.24} {\scriptstyle \pm 0.23}\)
           & \(0.37 {\scriptstyle \pm 0.10}\) & \(0.35 {\scriptstyle \pm 0.10}\) 
           & \(0.81 {\scriptstyle \pm 0.07}\) & \(0.81 {\scriptstyle \pm 0.08}\) 
           & \(3708 {\scriptstyle \pm 1243}\) & \(3473 {\scriptstyle \pm 1355}\) \\

DIAMOND-50 & \includegraphics[width=0.02\linewidth]{contents/tick.png}  
           & \(\mathbf{0.18} {\scriptstyle \pm 0.18}\) & \(\mathbf{0.22} {\scriptstyle \pm 0.23}\)
           & \(0.37 {\scriptstyle \pm 0.10}\) & \(0.38 {\scriptstyle \pm 0.09}\) 
           & \(0.83 {\scriptstyle \pm 0.09}\) & \(0.81 {\scriptstyle \pm 0.08}\) 
           & \(3249 {\scriptstyle \pm 833}\)  & \(2994 {\scriptstyle \pm 906}\) \\

\midrule

NWM-5   & \includegraphics[width=0.02\linewidth]{contents/tick.png}  
        & \(\mathbf{0.40} {\scriptstyle \pm 0.26}\) & \(\mathbf{0.39} {\scriptstyle \pm 0.30}\)
        & \(0.33 {\scriptstyle \pm 0.11}\) & \(0.31 {\scriptstyle \pm 0.09}\) 
        & \(0.64 {\scriptstyle \pm 0.05}\) & \(0.67 {\scriptstyle \pm 0.05}\) 
        & \(1950 {\scriptstyle \pm 380}\) & \(2240 {\scriptstyle \pm 664}\) \\

NWM-15  & \includegraphics[width=0.02\linewidth]{contents/tick.png}  
        & \(\mathbf{0.38} {\scriptstyle \pm 0.24}\) & \(\mathbf{0.40} {\scriptstyle \pm 0.25}\)
        & \(0.30 {\scriptstyle \pm 0.12}\) & \(0.33 {\scriptstyle \pm 0.12}\) 
        & \(0.67 {\scriptstyle \pm 0.03}\) & \(0.65 {\scriptstyle \pm 0.05}\) 
        & \(2132 {\scriptstyle \pm 916}\) & \(2338 {\scriptstyle \pm 1010}\) \\

NWM-30  & \includegraphics[width=0.02\linewidth]{contents/tick.png}  
        & \(\mathbf{0.38} {\scriptstyle \pm 0.21}\) & \(\mathbf{0.36} {\scriptstyle \pm 0.25}\)
        & \(0.32 {\scriptstyle \pm 0.11}\) & \(0.30 {\scriptstyle \pm 0.11}\) 
        & \(0.69 {\scriptstyle \pm 0.04}\) & \(0.71 {\scriptstyle \pm 0.03}\) 
        & \(1893 {\scriptstyle \pm 1047}\) & \(2437 {\scriptstyle \pm 429}\) \\

NWM-50  & \includegraphics[width=0.02\linewidth]{contents/tick.png}  
        & \(\mathbf{0.35} {\scriptstyle \pm 0.21}\) & \(\mathbf{0.39} {\scriptstyle \pm 0.22}\)
        & \(0.28 {\scriptstyle \pm 0.13}\) & \(0.33 {\scriptstyle \pm 0.11}\) 
        & \(0.72 {\scriptstyle \pm 0.08}\) & \(0.65 {\scriptstyle \pm 0.04}\) 
        & \(2715 {\scriptstyle \pm 883}\) & \(1537 {\scriptstyle \pm 415}\) \\

\bottomrule
\end{tabular}
\label{table}
\vspace{-10pt}
\end{table}

\section{Limitations and Future Work}
\label{limitation}
\textbf{Lack of depth data}: Many navigation and 3D modeling tasks require depth information to assist with memory modeling. While recent works like Depth Anything \citep{depthanything,depth_anything_v2} can generate synthetic depth data, there is still a gap between these synthetic depth maps and ground truth 3D data. We believe that real-time depth acquisition from the Minecraft server using scanning techniques could address this, but it would introduce substantial computational overhead.


\textbf{Static structures in the dataset}: The structures in our dataset (such as houses and terrain) are static, and we intentionally excluded moving objects like mobs. This decision was made to simplify the task of modeling spatial consistency. However, in real-world environments, the positions of vehicles, people, and other dynamic objects change over time. Maintaining spatial consistency in such dynamic environments is a more challenging task and a promising direction for future exploration. We leave this task for future work.

\section*{Acknowledgments}
This work was funded by the National University of Singapore under its Start-up Grant (Award No: SUG-251RES2505).
\bibliography{neurips_2026}
\bibliographystyle{plainnat}

\appendix

\section*{Appendix Overview}
Our appendix is organized as follows:
\begin{itemize}
    \item \textbf{Appendix~\ref{app:A-setup}} describes the platform setup for our data collection, including the Minecraft environment setup and three javascript plugins we modified to manipulate the bot and collect data.
    \item \textbf{Appendix~\ref{app:B-dataset}} presents detials of our dataset, including \ref{app:B2format} data storage format, \ref{app:B3stat} statistical information.
    \item  \textbf{Appendix~\ref{app:C-eval}} Evaluation Pipeline Details. \ref{app:C1-SGCS} introduces our \score metric.
    \item \textbf{Appendix~\ref{app:D-experiment}} presents the detailed experimental settings and additional experimental results:
    
    \textbf{\ref{app:D0-synthetic}}: Synthetic Perturbation Details. \textbf{\ref{app:D1-oasis}}: Open-oasis, \textbf{\ref{app:D2-mineworld}}: Mineworld, \textbf{\ref{app:D3-diamond}}: DIAMOND, \textbf{\ref{app:D4-nwm}}: Navigation World Model.
\end{itemize}

\section{Platform Setup}

\label{app:A-setup}
\begin{figure}[t!]
    \centering 
    \includegraphics[width=\textwidth]{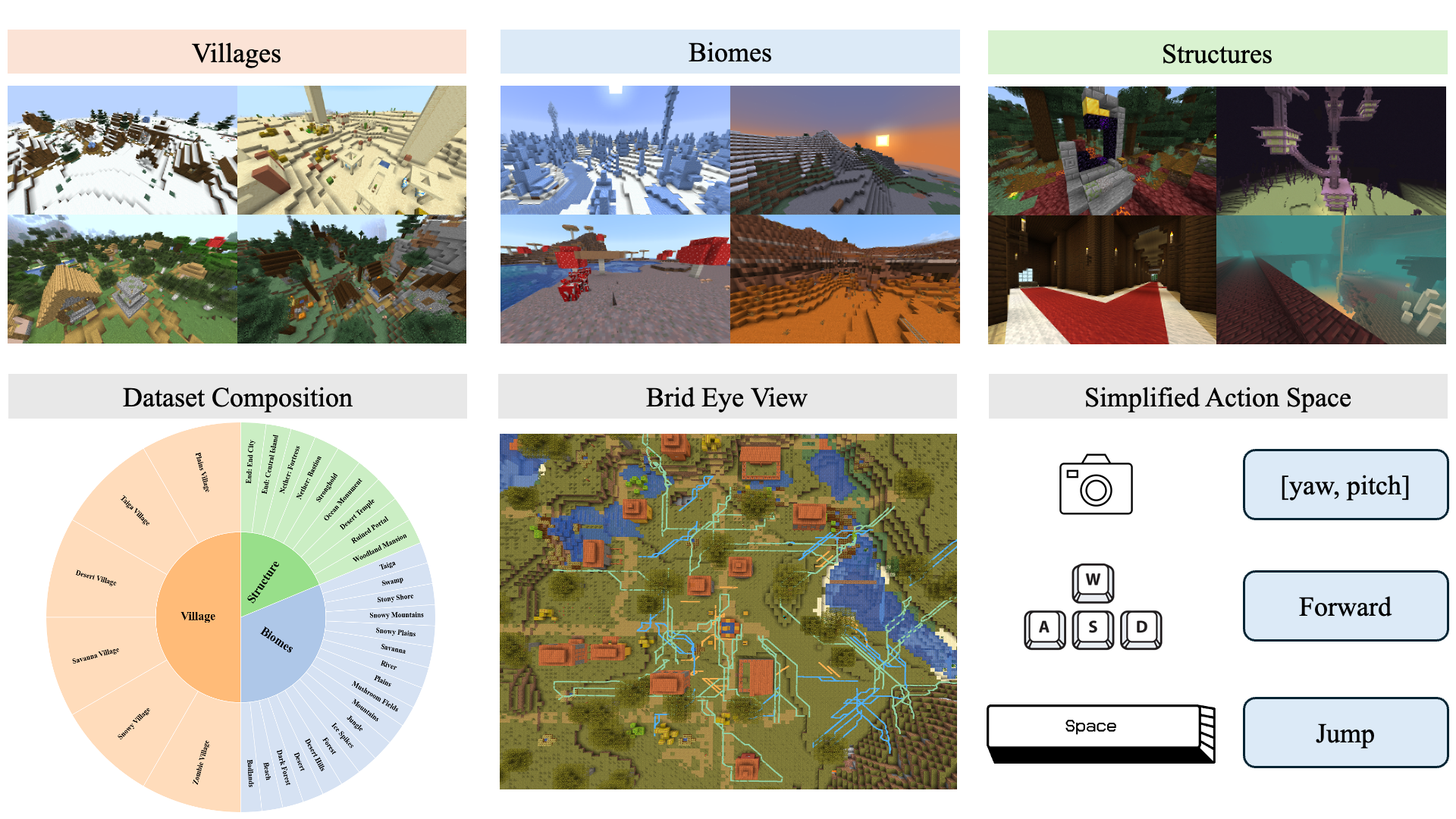}
    \vspace{-1.5em}
    \caption{\textbf{Overview of the Minecraft Elements}. Top row, left to right: Examples of villages, biomes, and structures in Minecraft. Bottom row, left to right: composition of sampling locations in our dataset; a bird-eye view of real ABA-type exploration trajectories (display steps 5, 15, and 30 for simplicity); Simplified Action Space used for data collection and agent interaction. }
    \label{fig:mc}
\end{figure}

We set up a local Minecraft server and use the Mineflayer platform to control the agent and collect trajectory data. For path planning, we employ the Mineflayer Pathfinder plugin, which computes shortest paths between waypoints using the A* algorithm. We use the Prismarine Viewer to render and visualize the agent’s behavior and collected observations.
To better support learning from the collected data and simulate human-like control dynamics, we make several key modifications to the default behavior of these plugins. These choices define some of the core characteristics of our dataset:

\textbf{Restricted action space}: The agent is limited to using only three actions during navigation — forward, jump, and camera rotation — closely matching the action primitives of a human player in Minecraft. See Figure \ref{fig:mc} for demostrations.

\textbf{Sequential action execution}: At any given time step, only a single action is allowed. That is, the agent cannot rotate the camera and move forward simultaneously. This avoids entangled motion patterns and improves the clarity of spatial transitions in the data.

\textbf{Smooth camera control}: We impose a maximum angular velocity on camera rotations to prevent abrupt viewpoint changes between frames, ensuring more stable visual continuity for training.

\textbf{Removal of irrelevant elements}: To focus the dataset on pure navigation, we disabled all entity spawning (e.g., mobs), and removed UI elements such as the hotbar and hands, eliminating visual distractions and ensuring task purity.

\subsection{Minecraft}
\label{app:A1-mc}
Minecraft is an open-world environment characterized by diverse terrains and rich interaction dynamics. We selected Minecraft for the following reasons:
\begin{itemize}
    \item \textbf{Rich environmental diversity}: Minecraft worlds are procedurally generated using random seeds, featuring a wide range of biomes and uniquely structured villages, enabling diverse and non-repetitive environments.
    \item \textbf{Extensive community resources}: Minecraft has a mature modding ecosystem that allows us to incorporate a wide variety of custom elements and tasks. In addition, there is a large amount of publicly available Minecraft gameplay footage on platforms like YouTube, which could be leveraged for future large-scale pretraining.
    \item \textbf{Rising research interest}: An increasing number of recent works have begun to study world modeling and agent learning in Minecraft, making it a timely and relevant platform for evaluating spatial consistency and memory mechanisms.
    \item  \textbf{Efficient simulation}: Compared to real-world data collection, Minecraft enables faster, cheaper, and highly parallelizable simulation, making it well-suited for large-scale controlled data generation.
\end{itemize}

We established a local Mineworld server running \textbf{Java Edition} version \textbf{1.16.5}, utilizing the seed value of \textbf{42}. The server was configured in survival mode with the difficulty set to peaceful, and mob spawning was disabled to eliminate the presence of animals. This configuration was implemented to prevent interference from mobs during trajectory-related experiments.

We utilized the Chunkbase website \footnote{http://www.chunkbase.com/} to locate various villages and biomes. A total of 120 villages were collected—comprising 6 types, with 20 samples each—along with locations of 18 distinct biomes and 8 structures. The agent was teleported to each initial position using Minecraft's teleport command before commencing navigation.

\subsection{Mineflayer}
Minefayer \footnote{https://github.com/PrismarineJS/mineflayer} is a powerful open-source JavaScript API designed for flexibly controlling in-game bots within Minecraft. It interacts with the Minecraft server by continuously reading block information around the bot based on its current x, y, and z coordinates, which is loaded locally. Users are allowed to specify actions for the bot to perform. Mineflayer simulates the bot’s movement physically, computes its resulting state and position after each action, and subsequently sends updated coordinates (x, y, z), yaw, and pitch values to the server. Benefiting from this flexible interaction mechanism, diverse action policies can be developed to enable the agent to accomplish various tasks.

We introduced a modification to Mineflayer by constraining the horizontal rotation speed per time step, such that each action can result in a \textbf{maximum rotation of 0.1 radians}. This adjustment ensures smoother and more gradual viewpoint transitions.

\subsection{Mineflayer-pathfinder}
The Mineflayer Pathfinder \footnote{https://github.com/PrismarineJS/mineflayer-pathfinder} is an application built on top of Mineflayer. It is capable of planning a path from the bot’s current position to a target coordinate (x, y, z) using the A* algorithm. The bot is then controlled through a series of actions to navigate toward the destination.

In the original implementation, camera rotation and movement were coupled within the same action. In our modified version, at any given time step, \textbf{only a single action is allowed}. That is, the agent cannot rotate the camera and move forward simultaneously. This avoids entangled motion patterns and improves the clarity of spatial transitions in the data.

Furthermore, the original Mineflayer Pathfinder often exhibited sharp turns and other non-smooth behaviors during navigation, particularly causing rapid camera jitter near jagged block edges. To mitigate this issue, we prevent the bot from moving along the edges of 1×1 blocks, thereby eliminating such undesirable motion artifacts.

\subsection{Prismarine-viewer}
Prismarine Viewer \footnote{https://github.com/PrismarineJS/prismarine-viewer} is a complementary plugin designed to work alongside Mineflayer, capable of rendering the Minecraft environment from either a first-person or third-person perspective. In this sense, the combination of Mineflayer and Prismarine Viewer effectively constitutes a functional Minecraft client. We employed Prismarine Viewer to render and generate gameplay visuals.

The rendering logic of the viewer operates by periodically capturing the current coordinates (x, y, z) yaw pitch and rendering the scene at a fixed time interval. We set this sampling frequency to 50 Hz, matching both Minecraft’s default refresh rate and the action-sending frequency. As a result, each frame is rendered every 0.02 seconds, synchronized with the execution of a new action.
\section{Dataset details and Visualization}
\label{app:B-dataset}
\subsection{Overview}
A total of 120 distinct villages were collected, comprising 6 types with 20 instances each. Our experiments were primarily conducted on this dataset of 120 villages. In addition, 18 different biomes and 8 types of locations were also gathered. Detailed information regarding the specific types of villages, biomes, and locations is provided in the table \ref{table:mc} below.
\begin{table}[h!]
\caption{Minecraft locations: Villages, Biomes, and Structures}
\label{table:mc}
\centering
\begin{tabular}{c|c c|c}

\textbf{Village} & \multicolumn{2}{c|}{\textbf{Biome}} & \textbf{Structure} \\
\hline
Plains Village   & Badlands         & Plains           & Woodland Mansion  \\
Savanna Village  & Beach            & River            & Ruined Portal     \\
Snowy Village    & Dark Forest      & Savanna          & Desert Temple     \\
Taiga Village    & Desert           & Snowy Plains     & Stronghold        \\
Desert Village   & Desert Hills     & Snowy Mountains  & Nether Bastion    \\
Forest Village   & Forest           & Stony Shore      & Nether Fortress   \\
Zombie Village   & Ice Spikes       & Swamp            & End Mainland      \\
                 & Jungle           & Taiga            & End City          \\
                 & Mushroom Fields  & Mountains        &                   \\
\end{tabular}
\end{table}

For each location, we collected two types of trajectories: \textbf{\{ABA, ABCA\}}. For each type, trajectories were generated with four different navigation ranges: \textbf{5, 15, 30, and 50}. We also collect \textbf{TURN} trajectories, which only contain camera rotation with angle \textbf{60-90,90-120,120-150,150-180}. Each length/angle comprises \textbf{20} distinct trajectories with varied start and end points.

This resulted in a total of 28,800 trajectories collected across all villages. The dataset contains around \textbf{15 million} frames, equivalent to roughly \textbf{200 hours} of gameplay.

For experimental purposes, the villages were partitioned into training, validation, and test sets as follows: villages 1-16 were used for training, villages 17–18 for validation, and villages 19–20 for testing. The trajectories were not shuffled randomly across splits to prevent the model from memorizing structural features seen during training. Instead, this partitioning strategy encourages the model to reconstruct environmental structures from contextual information, as each village has a unique layout.

\subsection{Data Format}
\label{app:B2format}
The smallest unit in our dataset consists of a pair of files: an \textbf{.avi video file} and a corresponding \textbf{.json file}, both of equal length. The .avi file stores the visual observations, while the .json file contains the associated state and action data. To avoid introducing inter-frame dependencies, we use MJPG (Motion JPEG) compression instead of H.264. Each frame is stored independently, and the resolution of the images is 640×360 at BGR format. The value of each BRG pixel is from [0,255]. The videos are recorded at 20 frames per second (FPS). This frequency aligns with the physical tick rate of Minecraft, ensuring consistent synchronization between control and simulation.

One step of recoreded trajectories in .json have following keys:

\begin{itemize}
    \item x: current x coordinate, rounded to three decimal places.
    \item y: current y coordinate, namely hight, rounded to three decimal places.
    \item z: current z coordinate, rounded to three decimal places.
    \item yaw: the agent’s horizontal viewing angle, measured in radians, ranging from $-\pi$ to $\pi$, where $0$ indicates the agent is facing the positive z-axis. 
    \item pitch: the agent’s vertical viewing angle, measured in radians, ranging from $-\pi/2$ to $\pi/2$, where 0 indicates the agent is looking straight ahead (parallel to the ground plane).
    \item action: a dictionary with three possible keys: forward, jump, and camera. The values of forward and jump are booleans indicating whether the corresponding action is executed. The camera key holds a tuple [ \text{yaw}, \text{pitch} ], representing the \textbf{change} in the agent’s viewing angle. Our sampling strategy ensures that these three actions do not occur simultaneously.
    \item goal: a dictionary containing the target’s coordinates, with keys x and z representing the target position on the x- and z-axes, respectively.
    \item frame count: the index of the current frame. Due to rendering initialization, the first 20 frames are skipped, so frame count starts at 20. Since we ensure that the entire trajectory is recorded properly, the actual frame index within the trajectory can be obtained by simply subtracting 20.
    \item extra info: seed, location, navigation type, navigation range.
\end{itemize}

\subsection{Statistical Information}
\label{app:B3stat}
\subsubsection{Trajectory Length}
To better illustrate the scale and difficulty gradient of the dataset, the distribution of trajectory lengths is presented in the Figure \ref{fig:frames} and Table \ref{table:length} below.

From the perspective of sequence length, even for the shortest navigation range (i.e., range = 5), the average trajectory length reaches as high as 180 frames. This implies that during evaluation, models must be capable of attending to visual inputs from over 100 frames ago. However, most current world models are limited to a history (context window) of only 32 frames, indicating that the required temporal context in our setting substantially exceeds the capacity of existing models.

In terms of distribution, different navigation ranges exhibit minimal overlap in their length distributions, which supports the validity of our difficulty curriculum design.

\begin{table}[!ht]
\caption{Mean Trajectory Frames ($\pm$ Std) for Different Trajectory Types}
\label{table:length}
\centering
\begin{tabular}{c|cccc}

\textbf{Trajectory Type} & \textbf{Range 5} & \textbf{Range 15} & \textbf{Range 30} & \textbf{Range 50} \\
\hline
ABA   & $180.5 \pm 30.4$   & $356.5 \pm 71.6$   & $627.1 \pm 130.3$   & $967.8 \pm 195.4$   \\
ABCA  & $251.0 \pm 47.1$   & $544.7 \pm 113.1$  & $968.1 \pm 203.7$   & $1362.6 \pm 260.1$  \\
\end{tabular}
\end{table}

\begin{figure}[ht]
    \centering 
    \includegraphics[width=\textwidth]{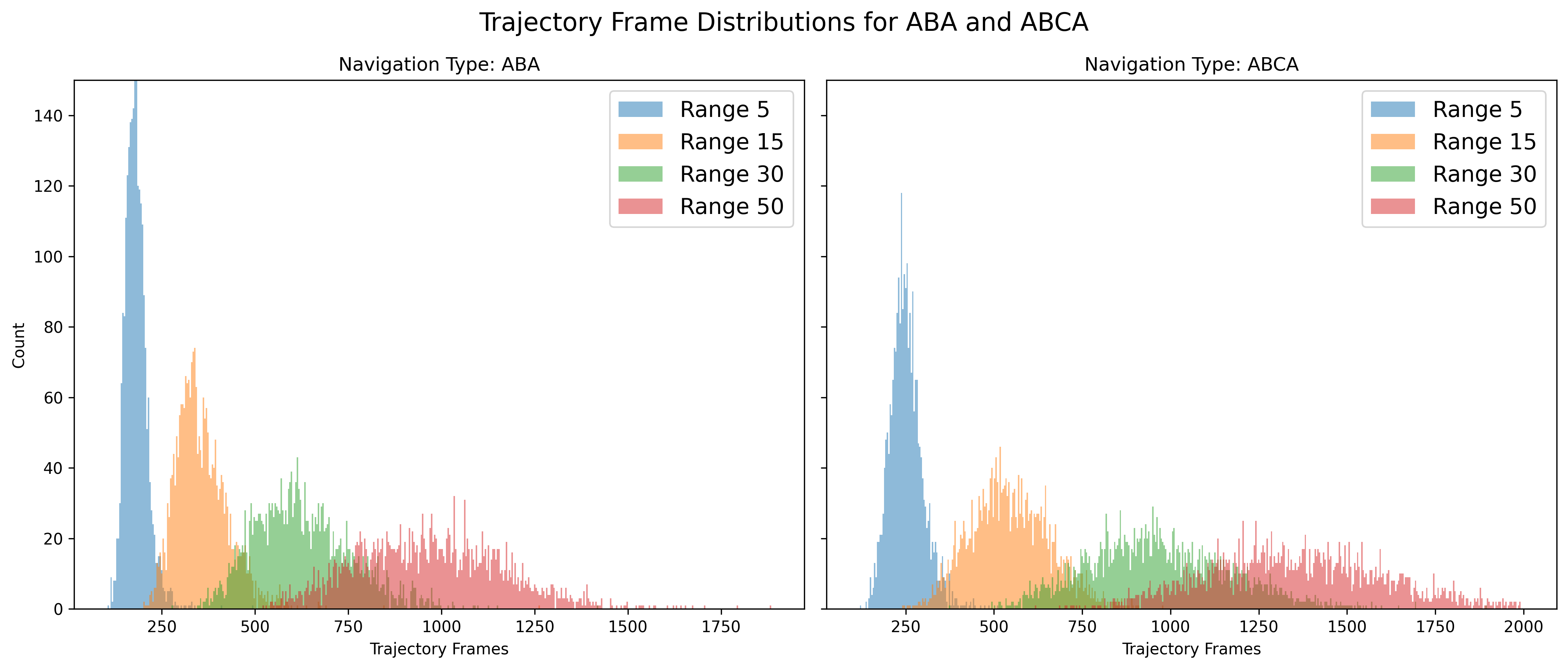}
    \vspace{-1.5em}
    \caption{\textbf{Trajectory Frame Distribution of Navgation Type ABA and ABCA}}
    \label{fig:frames}
\end{figure}

\subsubsection{Trajectory Variance}

In the original data collection pipeline, both the $A \to B$ and $B \to A$ paths were independently generated using A* search. Although this produced valid loop trajectories, the two paths could differ due to path-planning tie-breaking and local navigation artifacts. This introduces an additional source of trajectory variance that is unrelated to the intended spatial loop structure.

To reduce this confound, we revise the ABA collection protocol. The agent still uses A* search to navigate from $A$ to $B$, preserving the same start-point sampling, target-point sampling, navigation ranges, and environment settings as before. However, for the return trajectory, instead of running A* again from $B$ to $A$, the agent follows the recorded $A \to B$ waypoint sequence in reverse order. This reverse-waypoint return design encourages the $B \to A$ path to geometrically match the forward path as closely as possible, while still retaining physical execution, camera control, and rendering in the Minecraft environment.

We quantify the remaining discrepancy between the forward and return paths using the area enclosed by the loop. Let $\mathcal{A}$ denote the enclosed area and $L$ denote the total trajectory length of the forward and return paths. We report the normalized deviation as $2\mathcal{A}/L$, which corresponds to the average lateral deviation under a rectangular-loop approximation. The results are shown in Table~\ref{table:area}.

\begin{table}[h]
\centering
\caption{Deviation Between $A \to B$ and $B \to A$ Paths After Reverse-Waypoint Return}
\label{table:area}
\begin{tabular}{c|cccc}
\textbf{Metric} & \textbf{Range 5} & \textbf{Range 15} & \textbf{Range 30} & \textbf{Range 50} \\
\hline
Enclosed Area & $0.28 \pm 0.36$ & $0.54 \pm 1.08$ & $0.94 \pm 3.26$ & $1.66 \pm 10.72$ \\
Normalized Deviation & $0.046 \pm 0.041$ & $0.029 \pm 0.062$ & $0.025 \pm 0.074$ & $0.030 \pm 0.196$ \\
\end{tabular}
\end{table}

Compared with independently planning both directions using A*, the reverse-waypoint return substantially reduces the geometric discrepancy between the forward and return paths. The enclosed area remains close to zero across all navigation ranges, even for the longest range of 50 grids. This indicates that the revised ABA trajectories better isolate the intended loop-closure behavior: the return phase revisits the previously traversed route rather than introducing a new path generated by a second A* search. 

\section{Evaluation Pipeline Details}
\label{app:C-eval}

\subsection{\score Computation Details}
\label{app:C1-SGCS}

We describe the full evaluation pipeline used to compute the \matric (\score) between two videos. The implementation follows a frame-wise comparison protocol with object-level matching and category-wise aggregation.

\paragraph{Frame Sampling and Alignment.}
Given two videos $A$ and $B$, we first sample frames with a fixed stride $s$ (default $s=5$). For each sampled frame index $t$, we extract the corresponding frame pair $(I_t^A, I_t^B)$. If the two frames have different spatial resolutions, they are resized to a common resolution defined by the minimum height and width across the pair.

\paragraph{Object Segmentation via Text Prompts.}
For each frame, we extract object masks using a text-conditioned segmentation model (SAM3). We employ SAM 3.1 (released March 27, 2026) \footnote{https://github.com/facebookresearch/sam3.git}\footnote{https://huggingface.co/facebook/sam3.1}for scene graph consistency evaluation. Specifically, we define a fixed set of semantic categories (we use \textit{house, tree, flower, grass, water, sky, sand, dirt} for Minecraft), and for each category, we query the model to obtain a set of binary masks. Masks with low confidence (threshold $0.5$ by default) are discarded. This produces, for each category $c$, a set of masks $\mathcal{M}_c^A$ and $\mathcal{M}_c^B$ for the two frames.

\paragraph{Mask Representation.}
Each mask is converted into a compact geometric representation consisting of:
\begin{itemize}
    \item \textbf{Centroid:} computed as the mean of foreground pixel coordinates;
    \item \textbf{Area:} computed as the number of foreground pixels.
\end{itemize}
Masks with zero area are ignored. This representation allows efficient spatial comparison while remaining robust to shape irregularities.

\paragraph{Instance Matching.}
For each category $c$, we match instances across the two frames using centroid distance. Let $\{x_i^A\}$ and $\{x_j^B\}$ denote the centroids. We compute pairwise Euclidean distances and normalize them by the image diagonal $D = \sqrt{H^2 + W^2}$. A pair is considered a valid match if:
\[
\|x_i^A - x_j^B\|_2 < \tau \cdot D,
\]
where $\tau=0.1$ by default. We then perform bipartite matching using the Hungarian algorithm to maximize the number of valid matches.

\paragraph{Category-Level Score.}
For each category $c$, let $m$ and $n$ denote the number of detected instances in the two frames, and $p$ the number of matched pairs. We define the category-level consistency score as:
\[
S_c = \frac{2p}{m+n}
\]
This formulation corresponds to an instance-level F1 (Dice) score, penalizing both missing and mismatched objects.

\paragraph{Area-Weighted Aggregation.}
To account for object importance, we weight each category by its spatial extent:
\[
w_c = \frac{\text{area}_c^A + \text{area}_c^B}{2}.
\]
The frame-level \score is computed as:
\[
\text{SGCS}_t = \frac{\sum_c w_c S_c}{\sum_c w_c}.
\]

\paragraph{Video-Level Aggregation.}
Finally, we compute the video-level \score by averaging over all sampled frames:
\[
\text{SGCS}_{video} = \frac{1}{T} \sum_{t=1}^{T} \text{SGCS}_t.
\]
In dataset-level evaluation, we further report the mean and standard deviation across trajectories, where each trajectory-level score is obtained via temporal averaging.

\subsection{Traditonal Evaluation Metrics}

\label{app:C2-traditional}

\textbf{Fréchet Video Distance (FVD)} We adopt Fréchet Video Distance (FVD) \citep{fvd} as a primary metric. FVD computes the distance between the distribution of real and generated video features extracted by an Inflated 3D ConvNet (I3D) \citep{I3D}, capturing both spatial and temporal statistics at a high level.

\textbf{Learned Perceptual Image Patch Similarity (LPIPS)} We use LPIPS \citep{lpips} to evaluate the perceptual similarity between generated and ground truth frames. LPIPS leverages deep features from pretrained networks (e.g., VGG) to assess semantic-level differences, which aligns better with human judgment than pixel-wise metrics. It is robust to minor texture or color variations while remaining sensitive to object structure and layout.

\textbf{Structural Similarity Index Measure (SSIM)} SSIM is employed to complement LPIPS by quantifying structural fidelity and luminance similarity between generated and reference frames. While it is more sensitive to low-level details than LPIPS, it still emphasizes perceptual structure over exact pixel match.

\subsection{Context Coverage in ABCA Trajectories}
\label{app:C3-ABCA}
For longer $A \rightarrow B \rightarrow C \rightarrow A$ trajectories, we evaluate only the $C \rightarrow A$ segment, treating $A \rightarrow B \rightarrow C$ as the exploration context. It is worth noting that in these trajectories, the return path from $C \rightarrow A$ is not guaranteed to be fully covered by the earlier observations from exploration context. However, in practice, we constrain the exploration range such that the majority of $C \rightarrow A$ trajectories traverse areas that have been previously observed, albeit from different viewpoints or at different times. As a result, the model typically has access to sufficient contextual information to support the reconstruction of the return path, making our evaluation of spatial consistency meaningful.

\subsection{Extension of \matric}
\label{app:C4-sgcsextend}
we have designed and implemented a pilot hierarchical extension of SGCS:
\begin{itemize}
    \item SGCS (Layout Level): The original category-conditioned spatial matching to evaluate high-level scene layout and object placement.
    \item SGCS-S (Shape Level): For spatially matched regions, computes translation-normalized mask IoU to measure the fine-grained shape alignment of generated objects.
    \item SGCS-T (Texture Level): For pairs that pass a predefined shape-matching threshold, evaluates visual surface appearance using standard AlexNet LPIPS computed over tightly cropped common-foreground regions.
\end{itemize}

The central design principle is hierarchical evaluation. Evaluation at a finer granular level is meaningful only when the lower level achieves sufficient alignment. If the spatial layout is completely divergent, comparing shape or texture becomes meaningless. Thus, shape should be evaluated given similar layout (SGCS), and texture should be evaluated given both layout and shape consistency (SGCS-S).

It is worth noting that current state-of-the-art world models still struggle to achieve satisfactory performance even at the basic layout alignment stage (SGCS). Consequently, SGCS-S and SGCS-T serve as forward-looking evaluation standards that impose higher-level capability requirements for future world model development.

We conducted preliminary experiments to validate the sensitivity and orthogonality of this hierarchical extension:

\textbf{Shape Perturbations}: Modifying object shape while maintaining location leaves the original SGCS score largely unaffected, whereas SGCS-S successfully detects and reflects the shape deviation. For shape, a local notch removes a localized circular region from the object boundary, producing a contour defect while preserving the object category and centroid. Dilation expands the mask outward by adding boundary pixels, thereby increasing the object silhouette and area, whereas erosion removes boundary pixels and shrinks the silhouette. After each intervention, the mask is recentered to preserve its original position. Results are in Table \ref{tab:shape_perturbations}.

\begin{table}[htbp]
  \centering
  \caption{Comparison of SGCS and SGCS-Shape under Various Shape Perturbations}
  \label{tab:shape_perturbations}
  \begin{tabular}{lcc}
    \toprule
    \textbf{Shape Perturbations} & \textbf{SGCS} & \textbf{SGCS-Shape} \\
    \midrule
    Identity                     & 1.000                & 1.000               \\
    Erode 1 px                   & 1.000                & 0.819               \\
    Erode 2 px                   & 1.000                & 0.656               \\
    Dilate 1 px                  & 1.000                & 0.842               \\
    Dilate 2 px                  & 1.000                & 0.725               \\
    Local notch 25\%             & 1.000                & 0.833               \\
    \bottomrule
  \end{tabular}
\end{table}

\textbf{Texture Perturbations}: Altering surface texture while keeping location and shape intact shows no drop in SGCS or SGCS-S, while SGCS-T successfully captures the appearance shift. Hue shift rotates the hue values inside the foreground mask while largely preserving its spatial structure. Pixel shuffle randomly permutes foreground RGB pixels, approximately preserving the set of colors while destroying their spatial organization and local texture. Matched-color noise replaces foreground pixels with random values sampled using the original object’s per-channel color mean and variance, preserving coarse color statistics while disrupting local appearance. All appearance interventions keep the category, position, and mask exactly unchanged. Notice this is a distance so Identity get a score of 0. Results are in Table \ref{tab:texture_perturbations}.

\begin{table}[htbp]
  \centering
  \caption{Evaluation of Metric Sensitivity under Texture Perturbations}
  \label{tab:texture_perturbations}
  \begin{tabular}{lccc}
    \toprule
    \textbf{Texture Perturbations} & \textbf{SGCS} & \textbf{SGCS-Shape} & \textbf{SGCS-Texture (LPIPS $\downarrow$)} \\
    \midrule
    Identity                       & 1.000                & 1.000               & 0.000 \\
    Hue shift                      & 1.000                & 1.000               & 0.043 \\
    Pixel shuffle                  & 1.000                & 1.000               & 0.177 \\
    Matched-color noise            & 1.000                & 1.000               & 0.247 \\
    \bottomrule
  \end{tabular}
\end{table}

\section{Experiment Settings and Results}

\label{app:D-experiment}
We provide a detailed description of the experimental settings to facilitate reproduction.
All inference and training experiments were conducted on NVIDIA GeForce RTX 4090 GPUs, with a total compute time of approximately 1000 GPU hours respectively.
\subsection{Synthetic Perturbation Details}
\begin{figure}
    \centering
    \includegraphics[width=\linewidth]{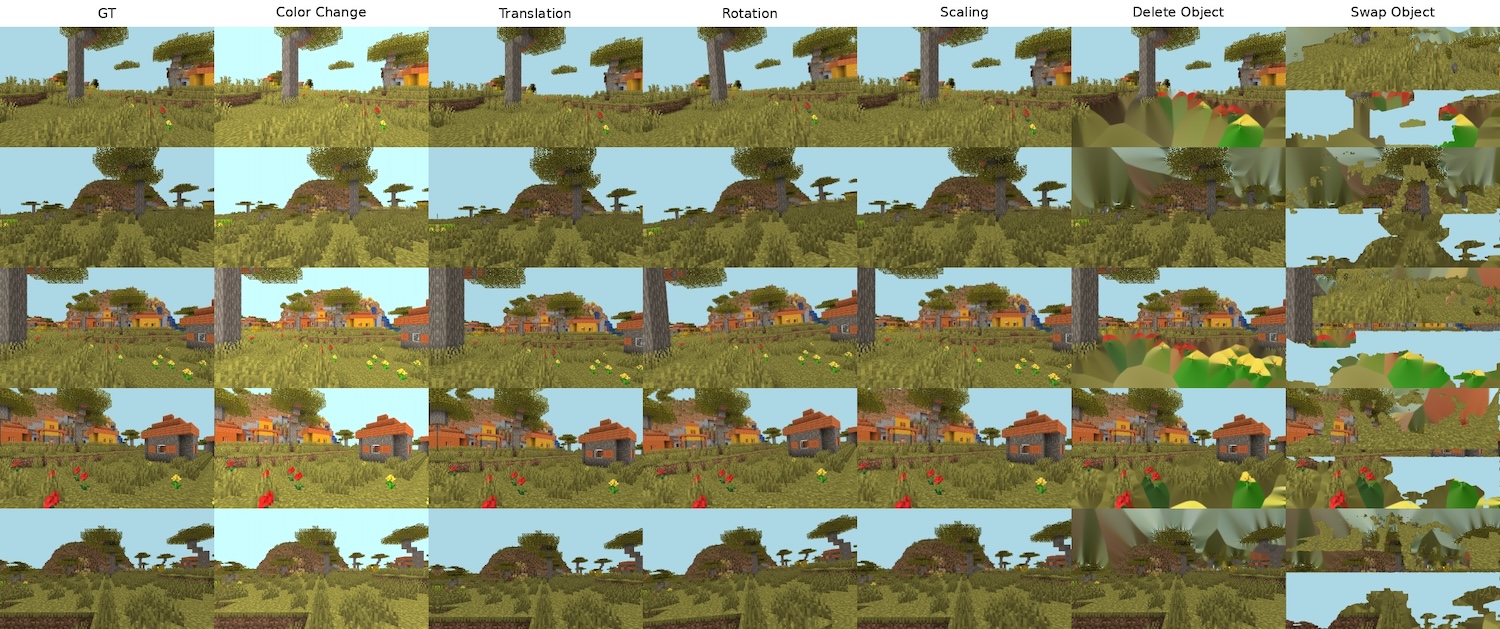}
    \caption{Synthetic Perturbation Demonstrations.}
    \label{fig:synthetic}
\end{figure}
\label{app:D0-synthetic}

This experiment is conducted on the navigation dataset with navigation type ABA and sequence length $5$, covering all $6$ distinct villages. Since our evaluation applies synthetic perturbations directly to ground-truth frames, the navigation type and trajectory length do not significantly affect the results. For reporting variability, we compute the standard deviation over video-level scores, where each video-level score is obtained by averaging frame-level metrics within a single video. We introduce all the synthetic perturbation details below, and the demonstration is in Figure \ref{fig:synthetic}.

\paragraph{Color change.}For the color-change perturbation, we modify only the appearance of the input frame while preserving its spatial layout. Specifically, each RGB frame is first adjusted by applying a contrast multiplier of $1.12$ and adding a brightness offset of $18.0$ to all pixel values. The result is clipped to the valid range $[0,255]$. We then convert the image to HSV space and further increase the saturation channel by a factor of $1.15$, followed by conversion back to RGB. This perturbation changes low-level color and illumination statistics without modifying object geometry or object layout.

\paragraph{Small translation.}For the small-translation perturbation, we apply a global 2D shift to the entire frame. The horizontal and vertical displacements are set to $4\%$ of the frame width and height, respectively. Formally, for a frame of size $W \times H$, the translation offsets are $\Delta x = 0.04W$ and $\Delta y = 0.04H$, rounded to integer pixel values with a minimum displacement of one pixel. The translated frame is generated using bilinear interpolation, and missing boundary regions are filled using reflection padding. This perturbation introduces pixel-level misalignment while preserving the relative spatial arrangement of objects.

\paragraph{Small rotation.}For the small-rotation perturbation, we rotate the entire frame around its image center by $5^\circ$ while keeping the image resolution unchanged. The transformation is implemented using an affine rotation matrix with scale fixed to $1.0$. Bilinear interpolation is used for resampling, and reflection padding is used to fill boundary regions. Since all objects undergo the same rigid transformation, the global image alignment changes, but the underlying scene structure and object relationships are largely preserved.

\paragraph{Scaling via crop-and-resize.}For the scaling perturbation, we simulate a zoom-in transformation using center crop followed by resizing. Specifically, we take a centered crop covering $90\%$ of the original frame height and width, and then resize the cropped region back to the original resolution. This produces a mild global scaling effect while maintaining the main scene layout. Because the transformation affects the whole frame consistently, it should not substantially alter object-level spatial consistency, although pixel-level similarity metrics may be affected by the resampling and scale change.

\paragraph{Object deletion.}For the object-deletion perturbation, we first obtain object masks using SAM with the predefined text categories. Candidate object masks are filtered by area: masks smaller than $0.2\%$ of the image area or larger than $40\%$ of the image area are discarded. Duplicate masks are removed using an IoU threshold of $0.85$. Among the remaining candidates, the largest object mask is selected for deletion. The selected mask is dilated with a $5 \times 5$ kernel and removed using Telea inpainting. This perturbation directly violates spatial consistency by removing a detected object from the scene while keeping the surrounding visual content plausible.

\paragraph{Object position swapping.}For the object-swapping perturbation, we again use SAM-derived object masks filtered with the same area constraints: minimum area fraction $0.002$, maximum area fraction $0.4$, and duplicate removal with IoU threshold $0.85$. If at least two valid objects are found, we select the two largest object candidates. The union of their masks is first removed from the image using inpainting. Then, each object patch is pasted into the other object's bounding box. During pasting, the source patch and source mask are resized to match the target bounding box, and only pixels inside the resized mask are copied. This perturbation preserves the presence of objects but changes their spatial locations, thereby creating a structural inconsistency in object layout.

\subsection{Open Oasis}
\label{app:D1-oasis}
\subsubsection{Oasis Settings}
We use the open-sourced Oasis-500M model for inference only. We do not train the Oasis model ourselves, as it is already pretrained on Minecraft VPT contractor data, and the official training code has not been released.
We use 32 frames as the conditioning input, wthich is the maximum supported context length, and perform inference in an auto-regressive manner.

Their observation space is defined as (640, 360, 3), which is fully consistent with our dataset and thus requires no additional modifications. Their action space follows VPT’s CameraQuantizer, with a maximum value of 20 and a bin size of 0.5, resulting in 40 discrete buckets that are normalized to the range [-1, 1]. In contrast, our actions are defined in radians within the range [-0.1, 0.1], and we directly use the raw values as actions. Additionally, their definitions of cameraX and cameraY are reversed compared to ours(and VPT's). Other actions, such as forward and jump, are consistent between the two settings. Inference hyperparameters are shown in Table \ref{table:oasis}
\begin{table}[ht]
\centering
\caption{Oasis Inference Hyperparameters}
\label{table:oasis}
\begin{tabular}{ccc}
\textbf{Model} & \textbf{Video Encoder} & \textbf{Sampling} \\
\hline
Model: \texttt{oasis-500M} & Model: ViT-VAE-L/20 & DDIM steps: 10 \\
Structure: DiT-S/2 & VAE patch size: 20 & Max noise level: 1000 \\
Model.max\_frames: 32 & \# Prompt frames: 32 & Noise absolute max: 20 \\
& & Stabilization level: 15 \\
& & Beta schedule: Sigmoid \\
\end{tabular}
\end{table}

\subsubsection{Oasis Results}
\begin{figure}[ht]
    \centering 
    \includegraphics[width=\textwidth]{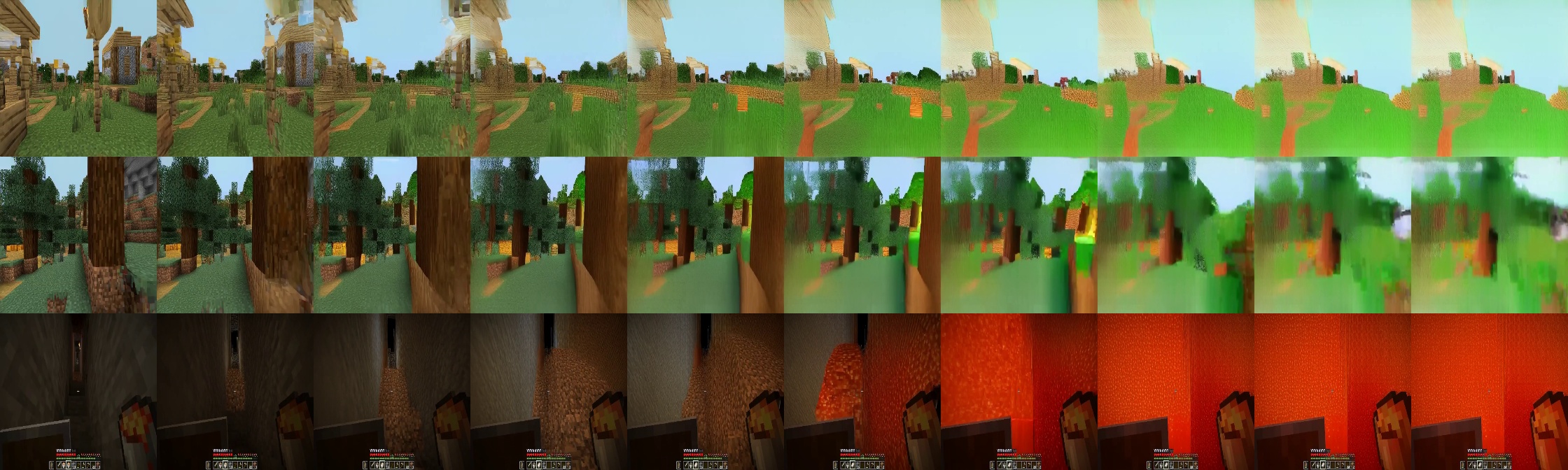}
    \vspace{-1.5em}
    \caption{Results of Oasis. The first row corresponds to plains village, the second row to taiga village, and the third row shows the official inference demo. All three cases exhibit increasing blurriness.}
    \label{fig:oasis}
\end{figure}
However, the results fall short of expectations. As shown in Figure \ref{fig:oasis}, we observe that errors compound over time: small imperfections quickly snowball into collapsed frames. After approximately 30 frames, the generated images become increasingly blurred and fail to recover. We include a sample from plains village and a sample from taiga village. To demonstrate that the blurriness is not caused by our dataset or action configuration, we additionally include the official inference demo (Player729-f153ac423f61-20210806-224813, 256 frames) in line 3 for comparison.

\subsection{Mineworld}
\label{app:D2-mineworld}
\subsubsection{Mineworld Settings}

Mineworld is also a model pretrained on Minecraft VPT contractor data, and its official training code has not been released. For the same reason, we directly evaluate Mineworld on our benchmark without further training. Specifically, we evaluate the largest publicly available model, Mineworld-1200M-16f and Mineworld-1200M-32f, which support 16 frames and 32 frames as context respectively.

Mineworld supports up to 32 frames of total context, which includes both historical and previously generated frames. Its generation process is chunk-based. We experiment using sliding window strategies: using 31 frames as context to autoregressively generate the remaining frames.

Mineworld defines its observation space as 384×224. To align with this, we resize our original 640×360 frames to 384×224 using an area-based interpolation method, and apply standard normalization to the pixel values. In Mineworld, visual observations are further compressed by a VAE into a latent representation of size 24×14. As for the action space, since Mineworld uses degrees to represent camera rotations, we convert our radian-based actions into degrees before passing them to the model.

\subsubsection{Mineworld Results}
\begin{figure}[ht]
    \centering 
    \includegraphics[width=\textwidth]{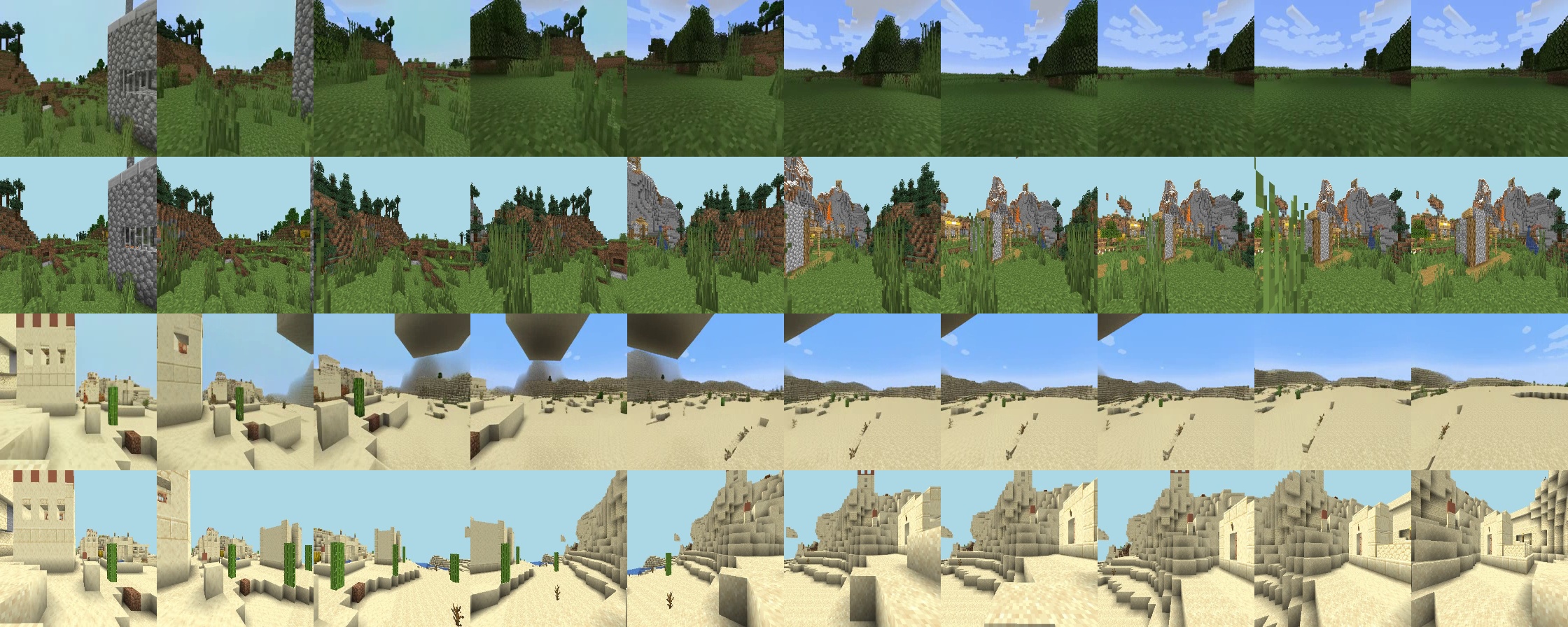}
    \vspace{-1.5em}
    \caption{Results of Mineworld.The first and second rows show the rollout and ground truth for plains village, respectively; the third and fourth rows show the same for desert village.}
    \label{fig:Mineworld}
\end{figure}

To qualitatively evaluate the model’s predictions, we visualize rollout results alongside ground truth frames in Figure \ref{fig:Mineworld}. The first and third rows show rollouts for plains village (village ID:20, trajectory ID: 05-10\_14-09-18) and desert village(village ID:20, trajectory ID: 05-10\_14-11-32), respectively, while the second and fourth rows present their corresponding ground truth trajectories. We observe that, unlike Oasis, Mineworld does not exhibit visual collapse. However, it similarly lacks spatial consistency and fails to reconstruct the corresponding scene structure.

\subsection{DIAMOND}
\label{app:D3-diamond}
\subsubsection{DIAMOND Settings}
DIAMOND \citep{diamond} is a diffusion-based world model built upon the UNet architecture \citep{unet}. It generates video frames autoregressively, conditioning on both past observations and actions, allowing it to model complex temporal dynamics in sequential decision-making environments. In its original design, DIAMOND is used as part of a model-based reinforcement learning pipeline, where the learned world model is used to train an RL agent and evaluated on the Atari 100k benchmark. And in CS:GO branch, DIAMOND also exhibits extraordinary ability to modeling dynamics. We follow the \texttt{CS:GO} configuration and focus solely on training the world model, omitting the reinforcement learning (RL) agent training phase. For the observation space, following the default settings of the CS:GO branch in DIAMOND, we first resize the input images from $640 \times 360$ to $320 \times 180$, and then further downscale them to $64\times36$. Diffusion is performed on the $64\times36$ images, and an additional upsampler is trained to reconstruct the images back to $320 \times 180$.

For the action space, we adopt VPT-style camera quantization with a maximum value of 0.1 and a bin size of 0.02, resulting in 11 discrete bins for both yaw and pitch. Combined with forward and jump actions, the total action dimension is 24. The training hyperparameters are summarized in Table \ref{table:diamond}.

\begin{table}[ht]
\centering
\caption{Model Configuration Parameters}
\label{table:diamond}
\begin{tabular}{lccc}
\hline
\textbf{Category} & \textbf{Parameter} & \textbf{Denoiser} & \textbf{Upsampler} \\
\hline
\multirow{4}{*}{General} & sigma\_data & 0.5 & 0.5 \\
 & sigma\_offset\_noise & 0.1 & 0.1 \\
 & noise\_previous\_obs & true & false \\
 & upsampling\_factor & null & 5 \\
\hline
\multirow{3}{*}{Inner Model} & img\_channels & 3 & 3 \\
 & num\_steps\_conditioning & 4 & 1 \\
 & cond\_channels & 2048 & 2048 \\
\hline
\multirow{9}{*}{Training} & num\_autoregressive\_steps & 4 & 1 \\
 & start\_after\_epochs & 0 & 0 \\
 & steps\_first\_epoch & 400 & 400 \\
 & steps\_per\_epoch & 400 & 400 \\
 & sample\_weights & null & null \\
 & batch\_size & 64 & 16 \\
 & grad\_acc\_steps & 2 & 2 \\
 & lr\_warmup\_steps & 100 & 100 \\
 & max\_grad\_norm & 10.0 & 10.0 \\
\hline
\multirow{3}{*}{Optimizer} & lr & 1e-4 & 1e-4 \\
 & weight\_decay & 1e-2 & 1e-2 \\
 & eps & 1e-8 & 1e-8 \\
\hline
\multirow{10}{*}{Diffusion Sampler}
 & num\_steps\_denoising & 3 & 10 \\
 & sigma\_min & 2e-3 & 1 \\
 & sigma\_max & 20.0 & 5.0 \\
 & rho & 7 & 7 \\
 & order & 1 & 1 \\
 & s\_churn & 0.0 & 10.0 \\
 & s\_tmin & 0.0 & 1 \\
 & s\_tmax & $\infty$ & 5 \\
 & s\_noise & 1.0 & 0.9 \\
 & s\_cond & 0.005 & 0 \\
\hline
\end{tabular}
\end{table}

\subsubsection{DIAMOND Results}
\begin{figure}[ht]
    \centering 
    \includegraphics[width=\textwidth]{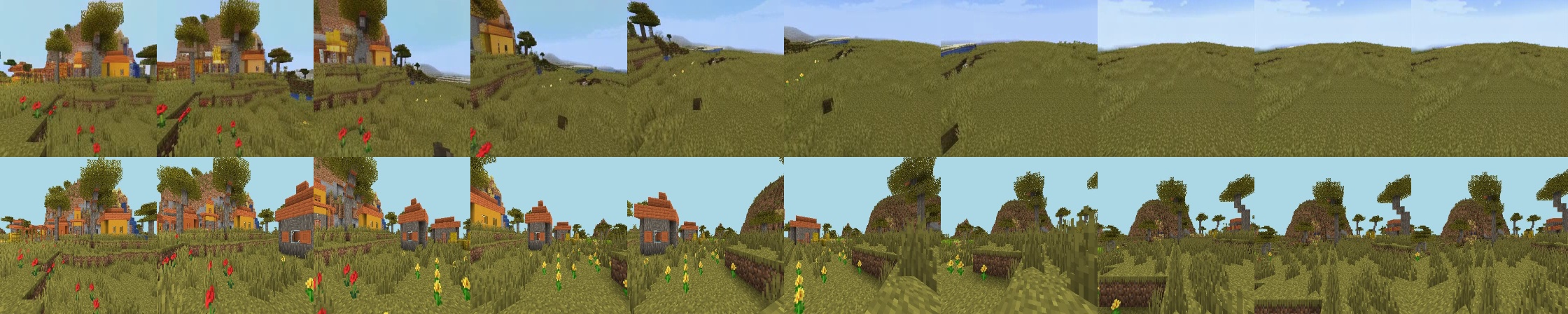}
    \vspace{-1.5em}
    \caption{Results of Diamond. First row: Rollout results. Second row: Ground truth.}
    \label{fig:dia}
\end{figure}

In Figure \ref{fig:dia}, we present the rollout results of DIAMOND. We observe that DIAMOND does not suffer from visual collapse and is able to reconstruct the initial frames relatively accurately, which aligns with its context window of length 32. However, as the rollout progresses, the generated scenes gradually converge to empty grasslands, indicating that the model forgets previously observed structures. This suggests that DIAMOND, like others, fails to maintain spatial consistency over long horizons.

\subsection{Navigation World Model}
\label{app:D4-nwm}
\subsubsection{NWM Settings}

Navigation World Model (NWM) proposes a world model based on the Conditioned Diffusion Transformer (CDiT). Regarding the observation space, we first align with NWM’s input format by decomposing trajectory videos into individual images. Following the official setting, each image is resized to 224×224, normalized, and then compressed into a 32×32 latent representation using the Stable Diffusion VAE.

For the action space, NWM differs from prior models in that it conditions on the agent’s $(x, z)$ position and yaw angle, using absolute location and orientation to reconstruct future observations. Since our dataset records the $(x, z)$ coordinates and yaw at each step, we can directly adopt this format. Note that pitch information is omitted here. Empirically, we find that for navigation tasks, pitch tends to be less critical than yaw.

We train a CDiT-L/2 model with a context window of 32 frames. Detailed training hyperparameters are provided in Table \ref{table:nwm}.
\begin{table}[ht]
\centering
\caption{NWM Training Configuration}
\label{table:nwm}
\begin{tabular}{ll}
\hline
\textbf{Parameter} & \textbf{Value} \\
\hline
Batch size & 8 \\
Number of workers & 12 \\
Model & CDiT-L/2 \\
Learning rate & $8 \times 10^{-5}$ \\
Normalize action space & True \\
Gradient clipping value & 10.0 \\
Context size & 32 \\
\hline
\multicolumn{2}{l}{\textit{Distance Prediction}} \\
Min distance category & $-64$ \\
Max distance category & $64$ \\
\hline
\multicolumn{2}{l}{\textit{Action Output}} \\
Predicted trajectory length & 64 \\
\hline
\multicolumn{2}{l}{\textit{Dataset Settings}} \\
Image size & 224 \\
Goals per observation & 4 \\
\hline
\end{tabular}
\end{table}

\subsubsection{NWM Results}
\begin{figure}[ht]
    \centering 
    \includegraphics[width=\textwidth]{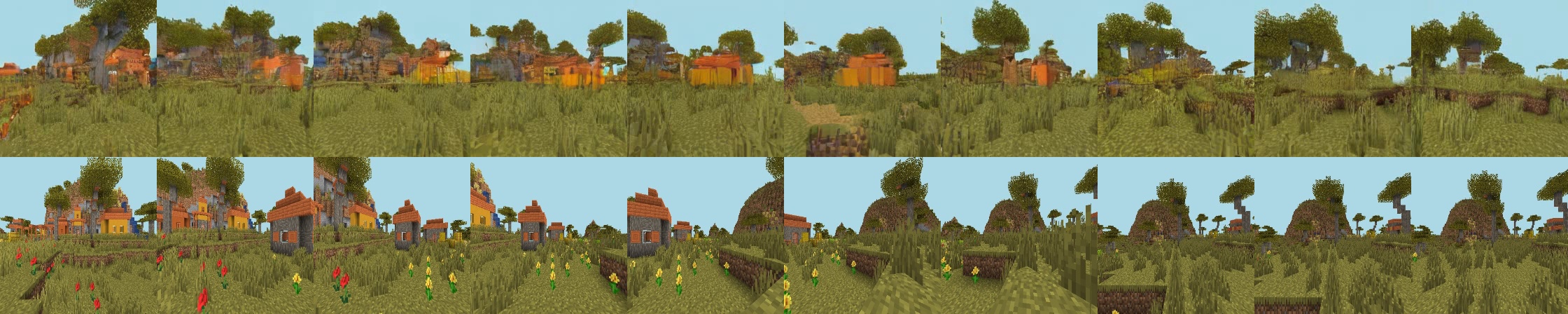}
    \vspace{-1.5em}
    \caption{Results of NWM. First row: Rollout results. Second row: Ground truth.}
    \label{fig:nwm}

\end{figure}

In Figure \ref{fig:nwm}, we present the rollout results of Navigation World Model. We observe that NWM retains a considerable amount of memory in the initial frames. However, the generated frames are noticeably blurry and distorted, suggesting that the model is unable to faithfully reconstruct spatial details or maintain long-range spatial consistency.

\subsection{Worldmem}
\subsubsection{WorldMem}
We initialize WorldMem from its released diffusion and VAE checkpoints.
LoopNav videos are sampled from 20 to WorldMem's native 10 FPS and resized
to \(640 \times 360\). We convert Mineflayer camera poses to WorldMem's
coordinate convention and derive continuous 25-dimensional actions from
the corrected pose changes. Each training sample contains eight local
frames and eight memory frames selected only from earlier observations.
We train on villages 1--16, validate on villages 17--18, and reserve
villages 19--20 for testing; TURN, ABA\_FIXED, and ABCA are sampled equally.
The VAE and base diffusion model remain frozen. We fine-tune the action
and pose conditioning modules and rank-8 LoRA adapters on the spatial,
temporal, and memory-attention QKV projections (18.5M trainable parameters)
for 20,000 optimizer steps using AdamW, a learning rate of
\(2 \times 10^{-5}\), and an effective batch size of 16. At evaluation,
WorldMem generates frames autoregressively at 10 FPS conditioned on the
recorded future actions and camera poses. We align predictions with the
corresponding source frames before scoring. 
\section{Extra Related Works}
\label{app:E-related}

\subsection{World Models}
The goal of a world model is to simulate the environment: Given the current state and action, it predicts the next state and the corresponding reward. World models were originally proposed to improve sample efficiency in reinforcement learning \citep{oh2015actionconditionalvideopredictionusing,worldmodelHa}. In the context of model-based reinforcement learning, numerous studies have explored various architectural designs of world models. Dreamer \citep{dreamer,dreamerV2,dreamerV3} uses a Recurrent State Space Model (RSSM) and achieves human-level performance on Atari games. TWM \citep{TWM} adapts DreamerV2’s RSSM to use a transformer architecture. IRIS \citep{iris} builds image tokens with discrete autoencoder and adopts an autoregressive transformer. DIAMOND \citep{diamond} leverages diffusion models to generate future frames. These architectures, to varying degrees, retain information from the past to aid in future image prediction; however, they still lack an effective memory design capable of maintaining long-horizon spatial consistency.

Beyond model-based reinforcement learning, the potential of world models has also been increasingly explored in other domains. Playable Video Generation \citep{menapace2021playablevideogeneration}, gameNGen \citep{gameNGen}, Oasis \citep{oasis} investigate the potential of using world models as game engines. GAIA \citep{hu2023gaia1generativeworldmodel}, DriveDreamer \cite{DriveDreamer}, vista \citep{vista} explores their application in autonomous driving. In robotics, two of the most significant challenges are the scarcity of training data and the high cost of model training. The emergence of world models offers a promising direction for addressing these two major challenges. UniSim \citep{unisim} learns a universal simulator for robot manipulation. DayDreamer \citep{wu2022daydreamerworldmodelsphysical} employs world model for real-world robot learning. NWM \citep{bar2024navigationworldmodels} investigates visual navigation problem in world model. In almost all downstream tasks—especially those involving decision-making such as autonomous driving and navigation—maintaining spatial consistency is a critical requirement.

From a more unified perspective, world models can be viewed as action-conditioned video generation models. In computer vision, generating videos has been a long standing challenge \citep{yang2024videonewlanguagerealworld}. Recent approaches leverage transformers, and diffusion models \citep{chen2024diffusionforcingnexttokenprediction,song2025historyguidedvideodiffusion} to generate longer, more coherent video sequences. From the perspective of video modeling, maintaining spatial consistency across frames—especially over long temporal horizons—remains a critical and unsolved challenge.

\subsection{Minecraft as an AI Testbed}
Minecraft is an open-world environment characterized by diverse terrains and rich interaction dynamics. As one of the most popular games globally, it benefits from extensive community-driven resources and an active user base. Recent research has increasingly adopted Minecraft as a platform for training generative agents \citep{vpt,steve1,cai2023groot,cai2024groot2weaklysupervisedmultimodal,cai2025rocket1masteringopenworldinteraction,cai2025rocket2steeringvisuomotorpolicy,pgt} and constructing digital world models \citep{guo2025mineworld,oasis,hong2024slowfastvgenslowfastlearningactiondriven,song2025historyguidedvideodiffusion}. 
Several Minecraft-based datasets and benchmarks have been introduced, including MineDoJo \citep{fan2022minedojo}, which provides multimodal knowledge and human gameplay videos sourced from YouTube; VPT \citep{vpt}, which releases large-scale trajectories of human players engaged in building and mining; and MineRL \citep{guss2019minerl}, which focuses on trajectories related to cave exploration. However, these efforts primarily emphasize prospective exploration and object interaction, with limited attention to the role of historical context. 
In contrast, our proposed dataset and benchmark are designed to promote and assess models’ abilities to leverage past observations and perform spatial reasoning. Our data collection pipeline is built on top of Mineflayer \citep{mineflayer}, a widely used rule-based Minecraft automation framework.

\newpage
\section*{NeurIPS Paper Checklist}

\begin{enumerate}

\item {\bf Claims}
    \item[] Question: Do the main claims made in the abstract and introduction accurately reflect the paper's contributions and scope?
    \item[] Answer: \answerYes{}{} 
    \item[] Justification: Our main claim is that spatial consistency is a critical yet under-addressed challenge in current world models. In this paper, we introduce a loop-based navigation dataset and benchmark specifically designed to evaluate and improve spatial consistency.

\item {\bf Limitations}
    \item[] Question: Does the paper discuss the limitations of the work performed by the authors?
    \item[] Answer: \answerYes{} 
    \item[] Justification: We include a part to discuss our limitation in Section \ref{limitation}.

\item {\bf Theory assumptions and proofs}
    \item[] Question: For each theoretical result, does the paper provide the full set of assumptions and a complete (and correct) proof?
    \item[] Answer: \answerNA{} 
    \item[] Justification: The paper does not include theoretical results.

    \item {\bf Experimental result reproducibility}
    \item[] Question: Does the paper fully disclose all the information needed to reproduce the main experimental results of the paper to the extent that it affects the main claims and/or conclusions of the paper (regardless of whether the code and data are provided or not)?
    \item[] Answer: \answerYes{} 
    \item[] Justification: We release our code and dataset in the link.

\item {\bf Open access to data and code}
    \item[] Question: Does the paper provide open access to the data and code, with sufficient instructions to faithfully reproduce the main experimental results, as described in supplemental material?
    \item[] Answer: \answerYes{} 
    \item[] Justification: We release our code and dataset in the link.

\item {\bf Experimental setting/details}
    \item[] Question: Does the paper specify all the training and test details (e.g., data splits, hyperparameters, how they were chosen, type of optimizer) necessary to understand the results?
    \item[] Answer: \answerYes{} 
    \item[] Justification: We include details of experiments in Section \ref{expsetting} and Appendix \ref{app:D-experiment}.

\item {\bf Experiment statistical significance}
    \item[] Question: Does the paper report error bars suitably and correctly defined or other appropriate information about the statistical significance of the experiments?
    \item[] Answer: \answerYes{} 
    \item[] Justification: We report std of the expriment results.

\item {\bf Experiments compute resources}
    \item[] Question: For each experiment, does the paper provide sufficient information on the computer resources (type of compute workers, memory, time of execution) needed to reproduce the experiments?
    \item[] Answer: \answerYes{} 
    \item[] Justification: We provide GPU useage in Appendix \ref{app:D-experiment}.
    
\item {\bf Code of ethics}
    \item[] Question: Does the research conducted in the paper conform, in every respect, with the NeurIPS Code of Ethics \url{https://neurips.cc/public/EthicsGuidelines}?
    \item[] Answer: \answerYes{} 
    \item[] Justification: Yes, the research presented in the paper fully complies with the NeurIPS Code of Ethics. 

\item {\bf Broader impacts}
    \item[] Question: Does the paper discuss both potential positive societal impacts and negative societal impacts of the work performed?
    \item[] Answer: \answerNA{} 
    \item[] Justification: There is no societal impact of the work performed.
    
\item {\bf Safeguards}
    \item[] Question: Does the paper describe safeguards that have been put in place for responsible release of data or models that have a high risk for misuse (e.g., pre-trained language models, image generators, or scraped datasets)?
    \item[] Answer: \answerNA{} 
    \item[] Justification: The datasets released in this paper do not present a high risk for misuse.

\item {\bf Licenses for existing assets}
    \item[] Question: Are the creators or original owners of assets (e.g., code, data, models), used in the paper, properly credited and are the license and terms of use explicitly mentioned and properly respected?
    \item[] Answer: \answerYes{} 
    \item[] Justification: Yes, all external assets used in this work—including code, datasets, and pretrained models—are properly credited. 

\item {\bf New assets}
    \item[] Question: Are new assets introduced in the paper well documented and is the documentation provided alongside the assets?
    \item[] Answer: \answerYes{} 
    \item[] Justification: Yes, all external assets used in this work—including code, datasets, and pretrained models—are properly credited.

\item {\bf Crowdsourcing and research with human subjects}
    \item[] Question: For crowdsourcing experiments and research with human subjects, does the paper include the full text of instructions given to participants and screenshots, if applicable, as well as details about compensation (if any)? 
    \item[] Answer: \answerNA{} 
    \item[] Justification: The paper does not involve crowdsourcing nor research with human subjects.

\item {\bf Institutional review board (IRB) approvals or equivalent for research with human subjects}
    \item[] Question: Does the paper describe potential risks incurred by study participants, whether such risks were disclosed to the subjects, and whether Institutional Review Board (IRB) approvals (or an equivalent approval/review based on the requirements of your country or institution) were obtained?
    \item[] Answer: \answerNA{} 
    \item[] Justification: The paper does not involve crowdsourcing nor research with human subjects.

\item {\bf Declaration of LLM usage}
    \item[] Question: Does the paper describe the usage of LLMs if it is an important, original, or non-standard component of the core methods in this research? Note that if the LLM is used only for writing, editing, or formatting purposes and does \emph{not} impact the core methodology, scientific rigor, or originality of the research, declaration is not required.
    \item[] Answer: \answerNA{} 
    \item[] Justification: Core method development in this research does not involve LLMs as any important, original, or non-standard components.

\end{enumerate}

\end{document}